\documentclass[]{bytedance_seed}

\usepackage[toc,page,header]{appendix}

\usepackage{CJKutf8}

\usepackage{xargs}  

\usepackage{todonotes}  

\usepackage{amsmath}
\usepackage{dsfont}

\usepackage{svg}

\usepackage{siunitx}
\usepackage{listings}
\usepackage{colortbl}
\usepackage{tabularx}
\usepackage{makecell}

\newcommand{\rrm}{RRM}

\newcommand{\gcpo}{GCPO}
\newcommand{\grpo}{GRPO}

\newcommand{\editrone}{Edit-R1}
\newcommand{\Ind}[1]{\mathds{1}\!\left\{#1\right\}}

\lstdefinestyle{promptstyle}{
  basicstyle=\ttfamily\footnotesize,
  breaklines=true,
  frame=single,
  frameround=tttt,
  rulecolor=\color{black!60},
  showstringspaces=false,
  numbers=left,
  numberstyle=\tiny\color{gray},
  stepnumber=1,
  numbersep=5pt,
  tabsize=2,
  captionpos=b,
}

\usepackage{amssymb}

\title{Leveraging Verifier-Based Reinforcement Learning in Image Editing}

\author{
Hanzhong Guo\texorpdfstring{$^{1,2}$}{} \quad
Jie Wu\texorpdfstring{$^{2,\dagger}$}{} \quad
Jie Liu\texorpdfstring{$^{2,4}$}{} \quad
Yu Gao\texorpdfstring{$^{2}$}{} \quad
Zilyu Ye\texorpdfstring{$^{2}$}{} \quad
Linxiao Yuan\texorpdfstring{$^{2}$}{} \quad
Xionghui Wang\texorpdfstring{$^{2}$}{} \quad
Yizhou Yu\texorpdfstring{$^{1,3}$}{}\texorpdfstring{$^{*}$}{} \quad
Weilin Huang\texorpdfstring{$^{2}$}{}\texorpdfstring{$^{*}$}{}
}

\affiliation[1]{School of Computing and Data Science, The University of Hong Kong}
\affiliation[2]{ByteDance Seed}
\affiliation[3]{Center for Embodied AI and Computer Vision, Shenzhen Loop Area Institute}
\affiliation[4]{CUHK}

\contribution[*]{Corresponding authors}
\contribution[\dagger]{Project lead}

\checkdata[Contact]{
\texttt{hanzhong@connect.hku.hk},
\texttt{wujie.10@bytedance.com},
\texttt{yizhouy@acm.org}
}

\abstract{
While Reinforcement Learning from Human Feedback (RLHF) has become a pivotal paradigm for text-to-image generation, its application to image editing remains largely unexplored.
A key bottleneck is the lack of a robust general reward model for all editing tasks.
Existing edit reward models usually give overall scores without detailed checks, ignoring different instruction requirements and causing biased rewards.
To address this, we argue that the key is to move from a simple scorer to a \textbf{reasoning verifier}. We introduce \textbf{\textit{Edit-R1}}, a framework that builds a chain-of-thought (CoT) verifier-based reasoning reward model (RRM) and then leverages it for downstream image editing. The Edit-RRM breaks instructions into distinct principles, evaluates the edited image against each principle, and aggregates these checks into an interpretable, fine-grained reward.
To build such an RRM, we first apply supervised fine-tuning (SFT) as a ``cold-start'' to generate CoT reward trajectories. Then, we introduce Group Contrastive Preference Optimization (GCPO), a reinforcement learning algorithm that leverages human pairwise preference data to reinforce our pointwise RRM.
After building the RRM, we use GRPO to train editing models with this non-differentiable yet powerful reward model.
Extensive experiments demonstrate that our Edit-RRM surpasses powerful VLMs such as Seed-1.5-VL and Seed-1.6-VL as an editing-specific reward model, and we observe a clear scaling trend, with performance consistently improving from 3B to 7B parameters. Moreover, Edit-R1 delivers gains to editing models like FLUX.1-kontext, highlighting its effectiveness in enhancing image editing.
}

\begin{document}
\begin{CJK*}{UTF8}{gbsn}

\maketitle

\begin{figure}[t]
\begin{center}
\vspace{-8pt}
\includegraphics[width=0.95\linewidth]{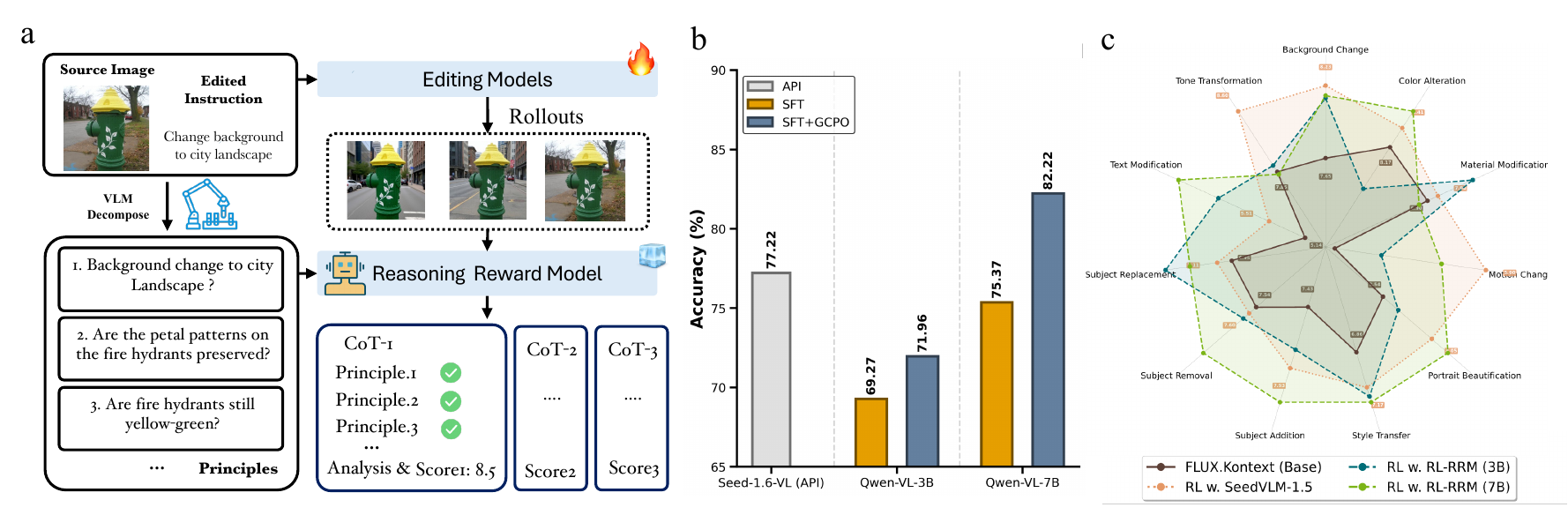}
\end{center}
\vspace{-6pt}
\caption{
\textbf{Our framework: from verifier-based reasoning reward model (RRM) to downstream.}
(a) \textbf{Verifier as a reasoning reward model.} The RRM decomposes an instruction into verifiable principles and scores an edited image against them in a single pass.
(b) \textbf{Reward benchmark performance.} Our final 7B model, trained with SFT and GCPO (RL-RRM), reaches 82.22\% accuracy, surpassing the Seed-VLM baseline. Each training component contributes to the performance gain.
(c) \textbf{Downstream application.} Using our 7B RL-RRM as a reward signal significantly improves the performance of FLUX.Kontext~\citep{batifol2025flux} across multiple editing categories during post-training.
}
\label{fig:r3l_pipeline_winloss}
\vspace{-8pt}
\end{figure}

\vspace{-1.8em}
\section{Introduction}
\label{sec:intro}

Image editing has evolved from earlier task-specific systems for photo adjustment and exemplar-based stylization~\cite{yan2016automatic,zhu2017exemplar,ruiz2023dreambooth,ye2023ip,guo2024real} to modern diffusion-based editors. 
With the rapid progress of diffusion and flow-based generative models, text-to-image (T2I) generation~\citep{ramesh2022hierarchical,saharia2022photorealistic,chang2023muse,balaji2022ediffi,rombach2022high,podell2023sdxl,esser2024scaling,betker2023improving,imagen2024imagen3,chen2023pixartalpha,chen2024pixartsigma,li2024hunyuandit,zheng2024cogview3,gong2025seedream,gao2025seedream,wu2025qwen,chen2025janus,han2025infinity}, image editing~\citep{brooks2023instructpix2pix,sheynin2024emu,ren2024byteedit,shi2024seededit,wang2025seededit,liu2025step1x,batifol2025flux,xiao2025omnigen}, and video generation~\citep{blattmann2023stable,bartal2024lumiere,brooks2024video,bao2024vidu,kong2024hunyuanvideo,moviegen2024movie,yang2025cogvideox,hacohen2025ltx,seawead2025seaweed,ma2025step,wan2025wan,gao2025seedance,seedance2025seedance,seedance2026seedance} have advanced dramatically.
In T2I generation, Reinforcement Learning from Human Feedback (RLHF) has become a core post-training step~\citep{gong2025seedream,gao2025seedream,wu2025qwen}, driven by powerful reward models (RMs)~\citep{xu2023imagereward,ma2025hpsv3,wang2025unified} and optimization algorithms~\citep{wallace2024diffusion,xu2023imagereward,xue2025dancegrpo}. By contrast, the application of RLHF to image editing has remained limited, with research still centered on pretraining and supervised fine-tuning (SFT)~\citep{wang2025seededit,deng2025emerging,batifol2025flux}.

A primary obstacle is the lack of a sufficiently robust reward model in editing. Image editing demands a more nuanced evaluation than T2I generation, assessing aspects like instruction fidelity, preservation of unedited regions, and overall quality. 
Existing approaches typically treat the RM as a holistic scorer, using a general-purpose Vision Language Model (VLM) to output a single, direct score~\citep{gong2025onereward,wei2025skywork}. 
This paradigm, also adopted by concurrent work like EditScore~\citep{luo2025editscore}, often fails to balance these complex aspects, leading to biased or hallucinated feedback~\citep{gunjal2025rubrics}. 
The key to overcoming this limitation is to shift from a simple "scorer" to a \textbf{reasoning verifier}—a model that explicitly decomposes the editing instruction, verifies the output against each sub-task, and then aggregates the results.

This paradigm shift presents two fundamental challenges.
 (1) \textit{Building and training a reliable verifier.} The first challenge lies in creating a verifier that can follow a structured reasoning process and accurately align with human preferences. While Chain-of-Thought (CoT) offers a promising structure, the initial training data is often noisy. Furthermore, aligning the verifier's complex, pointwise reasoning output with simple pairwise human preference data (e.g., "image A is better than B") is an unsolved problem, as standard RLHF algorithms like DPO~\cite{rafailov2023direct} or GRPO~\cite{guo2025deepseek} are ill-suited for this task.
 (2) \textit{RLHF Algorithms Compatible with RRMs}. While RLHF algorithms such as REFL have been applied to editing models~\citep{gong2025onereward,ren2024byteedit}, they are fundamentally incompatible with our reasoning reward model. Since the RRM generates an explicit multi-step reasoning trace through discrete token sampling before producing a final score, the process is inherently non-differentiable, rendering REFL-style methods inapplicable. Thus, a key challenge is how to leverage such a powerful RRM to achieve stable improvements in downstream editing models.

To address these challenges, we introduce \textbf{Edit-R1}, a framework for building and leveraging a verifier-based Reasoning Reward Model (RRM) to enhance image editing. Our approach begins by constructing the RRM to function as a verifier, evaluating edits by decomposing instructions into principles and generating a CoT analysis. To ensure high-quality "cold-start" training, we employ a powerful external VLM as a "quality-control judge" to filter the SFT data for the most plausible reasoning trajectories.

Subsequently, to align our pointwise RRM with pairwise human preference data, we introduce Group Contrastive Preference Optimization (GCPO), a novel reinforcement learning algorithm. 
GCPO is specifically designed to refine the RRM's reasoning capabilities by contrasting groups of "winner" and "loser" reasoning trajectories generated by the RRM itself. Once trained, this powerful, non-differentiable RRM serves as the verifier in a GRPO-based~\citep{xue2025dancegrpo, liu2025flow} reinforcement learning loop to significantly improve downstream editing models. Our contributions are summarized as follows:

\begin{itemize}[leftmargin=*]
    \item \textbf{A Verifier-based Reasoning Reward Model.} We propose a paradigm shift for image editing reward modeling, moving from a holistic scorer to a reasoning verifier. Our RRM, enabled by principle decomposition and CoT, provides more structured, interpretable, and reliable feedback.
    
    \item \textbf{A Novel RL Algorithm for RRM Training.} We introduce Group Contrastive Preference Optimization (GCPO), a new algorithm designed to optimize a pointwise, reasoning-based reward model using pairwise preference data. This method effectively refines the RRM's alignment with human judgments.
    
    \item \textbf{Superior Performance and Downstream Impact.} Our 7B RL-RRM significantly surpasses other RMs, including the concurrent EditScore~\citep{luo2025editscore}, on EditRewardBench~\citep{luo2025editscore}. When applied downstream, Edit-R1 delivers substantial gains to SOTA editors like FLUX.1-kontext and Qwen-Image-Edit, demonstrating the real-world effectiveness of our verifier-based RL framework.
\end{itemize}

\begin{table*}[t]
\centering
\resizebox{\textwidth}{!}{%
\begin{tabular}{l|c|c|c|ccc}
\toprule
\multirow{2}{*}{\textbf{Method}} & \multirow{2}{*}{\textbf{Task}}  & \multirow{2}{*}{\textbf{Modeling Paradigm}} & \multirow{2}{*}{\textbf{Point-wise}} & \multicolumn{3}{c}{\textbf{Reasoning Ability}} \\
& & & &  \textbf{As Verifier} & \textbf{With thinks} & \textbf{learned via RL} \\
\midrule
ImageReward~\citep{xu2023imagereward}          & Visual: T2I         &  Regressive  & $\checkmark$ & $\times$     & $\times$      & $\times$ \\
VideoAlign~\citep{liu2025improving}           & Visual: T2I           &  Regressive  & $\checkmark$ & $\times$     & $\times$      & $\times$  \\
WorldPM~\citep{wang2025worldpm}               & Understanding  &  Regressive & $\checkmark$ & $\times$     & $\times$      & $\times$ \\
DeepSeek-GRM~\citep{liu2025inference}         & Understanding  &  Generative  & $\times$     & $\checkmark$ & $\checkmark$  & $\checkmark$     \\
Pairwise RM~\citep{xu2025unified}             & Understanding  &  Generative  & $\times$     & $\times$     & $\checkmark$  & $\checkmark$ \\
UnifiedReward~\citep{wang2025unified,wang2025pref} & Multimodal &  Generative  & $\times$     & $\checkmark$ & $\checkmark$  & $\checkmark$     \\
RewardDance  ~\citep{wu2025rewarddance}                                & Visual: T2I          &  Generative  & $\times$     & $\times$     & $\checkmark$  & $\times$\\
OneReward~\citep{gong2025onereward}           & Visual: Edit         &  Generative  & $\times$     & $\times$     & $\times$  & $\times$     \\
VisualQuality-R1~\citep{wu2025visualquality}         & Visual: T2I          &  Generative  & $\checkmark$ & $\times$     & $\checkmark$  & $\checkmark$     \\
Skywork-EditReward~\citep{wei2025skywork}      & Visual: Edit            &  Generative  & $\checkmark$ & $\times$     & $\checkmark$  & $\times$     \\
EditScore~\citep{luo2025editscore}         & Visual: Edit     & Generative  &$\checkmark$     & $\times$     & $\times$  & $\times$     \\
\textbf{Edit-RRM (Ours)}                      & \textbf{Visual: Edit   } &  \textbf{Generative} & $\checkmark$ & $\checkmark$ & $\checkmark$  & $\checkmark$    \\
\bottomrule
\end{tabular}%
}
\caption{Comparison of reward models, highlighting reasoning capabilities. We categorize methods by their foundational characteristics (Task, Modeling Paradigm, etc.) and their support for advanced \textbf{Reasoning Ability} components: explicit use of principles(``as verifier''), Chain-of-Thought (``thinks''), and reinforcement learning. A checkmark ($\checkmark$) denotes support. \textbf{Edit-RRM (Ours)} is unique in integrating all three reasoning-enhancing features within a generative, point-wise framework for visual tasks.}
\label{tab:main}
\end{table*}
\section{Related Works}
\label{sec:realted}

\subsection{Reward model for generative models}
Driven by advances in Large Language Models (LLMs), many Reward Models (RMs) are now constructed directly upon them as shown in Tab.~\ref{tab:main}~\citep{ren2024byteedit,wu2025rewarddance,gong2025seedream,ma2025hpsv3}. 
In terms of modeling architecture, two dominant approaches have emerged, including regression-based~\citep{liu2025improving, wang2025worldpm} and generative-based~\citep{wu2025rewarddance,gong2025onereward,hong2025think}. The regression-based methods add a regression head for scoring, while the generative methods leverage the model's own generative abilities for assessment and are generally considered more effective at harnessing the base model's power. 
Regarding input format, methods are either pointwise~\citep{wu2025visualquality,wei2025skywork} or pairwise~\citep{wang2025unified, wu2025rewarddance}. Pointwise methods score a single response independently, while pairwise methods compare two responses to determine a preference. 
A significant drawback of pairwise approaches is their inability to provide an absolute quality score for a single response, making them ill-suited for direct quality assessment or filtering.
To enhance interpretability and accuracy, recent work has begun integrating Chain-of-Thought (CoT) reasoning and explicit principles~\citep{liu2025inference, wang2025unified}. For instance, DeepSeek-GRM~\citep{liu2025inference} utilizes principle-based CoT for generalist reward modeling. However, these methods are often designed for non-visual tasks. 
Our Edit-RRM including SFT-RRM and RL-RRM is uniquely positioned as a generative, pointwise verifier for image editing. It is the first to combine a principle-decomposition-based CoT process tailored for visual edits with undergoing a two-stage training pipeline, cleverly combining a rationale-based cold-start phase with subsequent reinforcement learning optimization.

\subsection{Reinforcement Learning in Image Editing}
Recent advancements in Reinforcement Learning from Human Feedback (RLHF) algorithms have demonstrated remarkable efficacy in aligning models with human preferences in the domain of image editing. 
DreamFuse~\citep{huang2025dreamfuse} adopts Direct Preference Optimization (DPO)~\citep{wallace2024diffusion} as their optimization method. 
However, DPO's direct optimization on a preference dataset inherently restricts policy exploration, risking suboptimal convergence.
While methods~\citep{gong2025onereward,ren2024byteedit} utilize REFL~\citep{xu2023imagereward} for preference alignment, REFL is often prone to severe reward hacking and requires the reward model to be differentiable.
Inspired by the notable success of DeepSeek-R1~\citep{guo2025deepseek}, many recent works~\citep{xue2025dancegrpo, liu2025flow} are now exploring the application of GRPO within the domain of visual generation. 
A key factor in DeepSeek-R1’s success was its reinforcement learning framework with verifiable rewards, which ensured robust training and mitigated the risk of reward hacking. Yet, defining such rewards for visual generation remains challenging. To address this, we extend the visual GRPO algorithm with a reasoning-based reward model for image editing, offering structural and principle-driven feedback.

\section{Method}

Proposed Edit-R1 introduces a framework centered around a \textbf{verifier-based Reasoning Reward Model (RRM)}. 
The core idea is to train this RRM to act as a reliable verifier and then leverage it to optimize downstream editing models. 
As detailed in Fig.~\ref{fig:mainfig_v2}, the training of our RRM is a two-stage process. 
Stage 1 (Cold-start SFT) constructs a large-scale, editing-specific SFT dataset. This stage focuses on quality, using an external VLM as a "quality-control judge" to select the most plausible "think+score" CoT trajectories from a diverse pool.
Stage 2 (GCPO) further refines the RRM using human preference pairs. For this, we introduce our novel Group Contrastive Preference Optimization (GCPO) algorithm, specifically designed to optimize a pointwise reasoning model with pairwise data. 
Finally, we integrate this powerful, non-differentiable RRM with the standard GRPO algorithm to elevate the performance of downstream editing models across multiple dimensions.


\subsection{Reward model}

\subsubsection{Verifier-based Reasoning Reward Model with Cold-Start}
\label{sec:cold_start_reward}

\begin{figure*}[t]
  \centering
  \includegraphics[width=\textwidth]{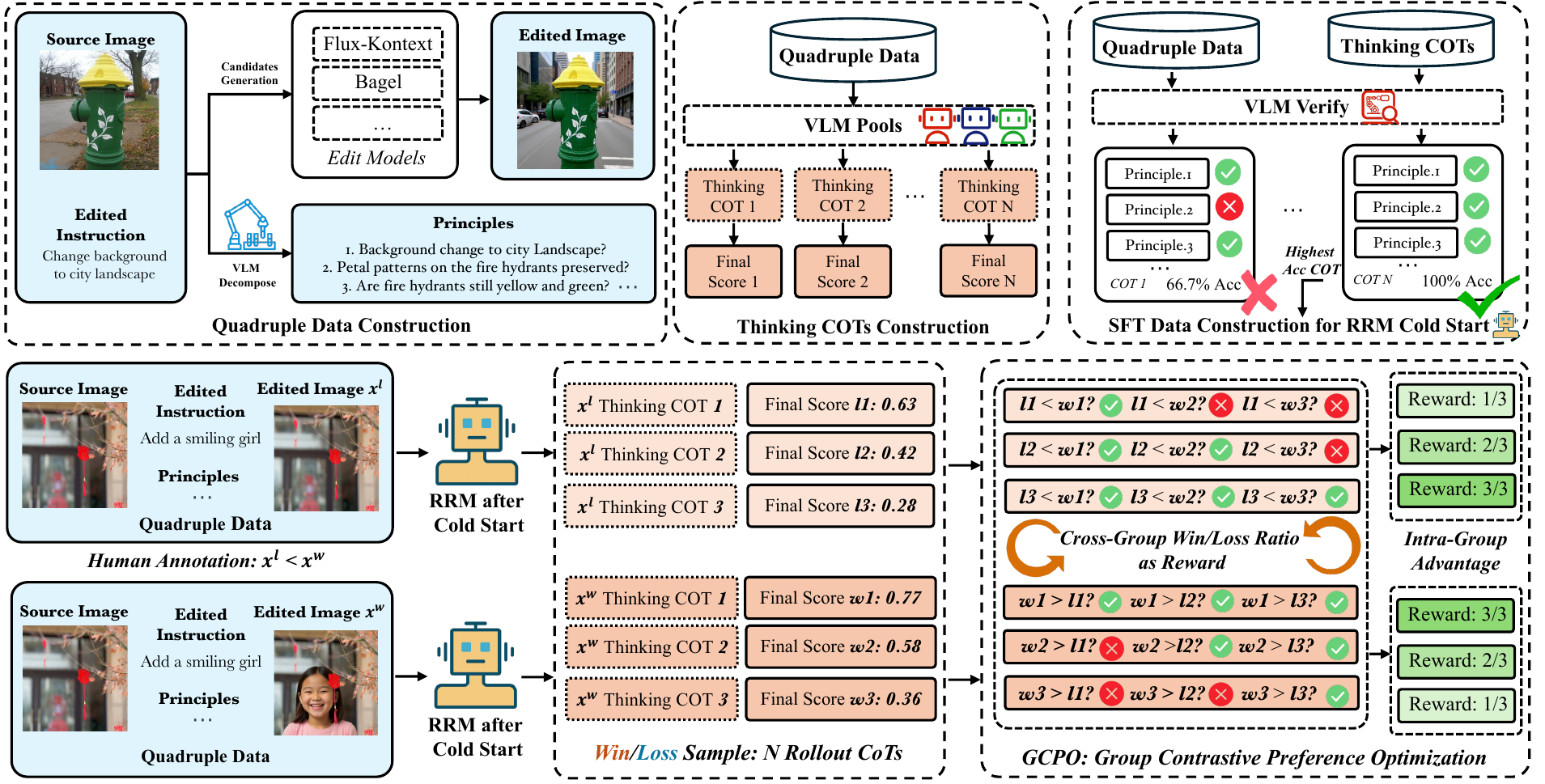}
  \caption{
  \textbf{The Training pipeline of Verifier-based Reasoning Reward Model (RRM).}
  \textbf{Top (Cold-Start SFT):} Given an edit instruction and a source image, we generate large-scale quadruple data (instruction, source image, principles, edited image) and employ VLM pools to generate numerous reasoning traces and use another VLM to select the thinking COT with the highest accuracy to build SFT data and cold-start the Reasoning Reward Model (RRM).
  \textbf{Bottom (GCPO):} For each human-labeled preference pair, the reward model generates $N$ thinking-score candidates per image. 
  We compute a win/loss ratio reward by pairwise comparing every candidate in the preferred group against all candidates in the non-preferred group. 
 The win ratio of a preferred candidate equals the fraction of comparisons where its score is \emph{higher} than the opposite group's scores;  The loss ratio of a non-preferred candidate equals the fraction where its score is \emph{lower} than the preferred group's scores. 
The advantage is computed within each preferred or non-preferred group.
  }
  \label{fig:mainfig_v2}
    \vspace{-12pt}
    
\end{figure*}

The goal of the first stage is to build a high-quality supervised dataset to "cold-start" our RRM. We begin by curating 200K samples from a public image-editing benchmark, partitioned into "Random" and "Hard" subsets to ensure diversity and complexity. 
i) Random Subset: The first 100K samples are randomly selected from the benchmark to represent a general distribution of edits. 
ii) Hard Subset: The second 100K samples are specifically curated for higher complexity. To achieve this, we employ GPT-4o to filter the remaining data and select edit instructions that require multi-step visual modifications, fine-grained detail editing, implicit semantic understanding, or precise spatial control, while rejecting simple single-step edits.
Each sample consists of a reference image $x_{\mathrm{ref}}$ and a corresponding edit instruction $q$.
The process, illustrated in the top panel of Fig.~\ref{fig:mainfig_v2}, follows four steps:
\vspace{-1.0em}
\paragraph{Step 1: Decomposing Instructions into Principles.}
For each reference image and its corresponding edit instruction, we employ the Seed-1.5-VL API to decompose the task into a concise set of verifiable principles using the system prompt in Appendix~\ref{sec:appendix_prompt}. These principles span three core aspects of image editing:
(a) Keep: elements that should remain unchanged;
(b) Follow: modifications required to align with the instruction;
(c) Quality: maintenance of generic visual integrity and fidelity.
This sample-wise decomposition effectively factorizes the editing task, structuring the model's reasoning process to distinguish between what to preserve and what to modify based on the specific input.
Formally, we denote the principle set as $\mathcal{P}=\{p_k\}_{k=1}^{K}$ for each (reference image, instruction) pair.
A concrete example of the decomposition is provided in Appendix~\ref{append:rrm}.
\vspace{-1.0em}
\paragraph{Step 2: Large-Scale Quadruple Generation.}
For each reference image and corresponding edit instruction, a diverse set of edited candidates is generated using multiple image-editing models, such as Flux-Kontext~\citep{blackforestlabs_flux}, Bagel~\citep{deng2025emerging}, and SeedEdit3.0~\citep{wang2025seededit}. Each edited candidate $x_{\mathrm{edit}}$, together with the reference input image $x_{\mathrm{ref}}$, the instruction $q$, and the principle set $\mathcal{P}$, forms a \emph{quadruple} $(x_{\mathrm{edit}}, x_{\mathrm{ref}}, q, \mathcal{P})$. This process yields a total dataset of approximately 2 million quadruples.

\vspace{-1.0em}
\paragraph{Step 3: VLM Reasoning and Point-wise Scoring.}
Each quadruple is processed by Vision-Language Models (VLM Pools) employing Chain-of-Thought (CoT) prompting. The VLM first performs a point-wise verification, assessing the edited image against each principle in $\mathcal{P}$. Subsequently, it generates a final scalar score representing the overall quality of the edited image, using the system prompt in Appendix~\ref{sec:appendix_prompt_rrm}. This score is computed as a weighted aggregate of the principle-wise verification outcomes. For more calculation methods and details, please refer to the appendix.
To enhance dataset diversity, we sample multiple thinking CoTs for each quadruple by varying system prompts, sampling temperatures, and VLM variants (e.g., Seed-1.5VL-1.5/Seed-1.6-VL), thereby producing multiple ``Think + Score" candidates.
We require the VLMs to generate the reasoning trace in a fixed format, verify principles in JSON format, and output the final score as \texttt{<score>...</score>}; a demo is shown in Appendix~\ref{append:rrm}.

\paragraph{Step 4: External Verification and SFT Data Selection.}
All ``Think + Score" candidates corresponding to the same quadruple are subjected to an external verification process. This is performed by SeedVLM-1.5, which functions as a \emph{point-wise verifier}. The verifier re-evaluates each principle in $\mathcal{P}$ for every reasoning trace and calculates a verification accuracy via the system prompt in Appendix~\ref{sec:verification}. We then select the thinking CoT that achieves the highest accuracy. The resulting data, comprising the instruction, images, principles, CoT reasoning trace, and final score, constitute the initial Supervised Fine-Tuning (SFT) dataset for the reward model's cold start.

\begin{figure*}[t!]
    \centering
    \includegraphics[width=0.98\textwidth]{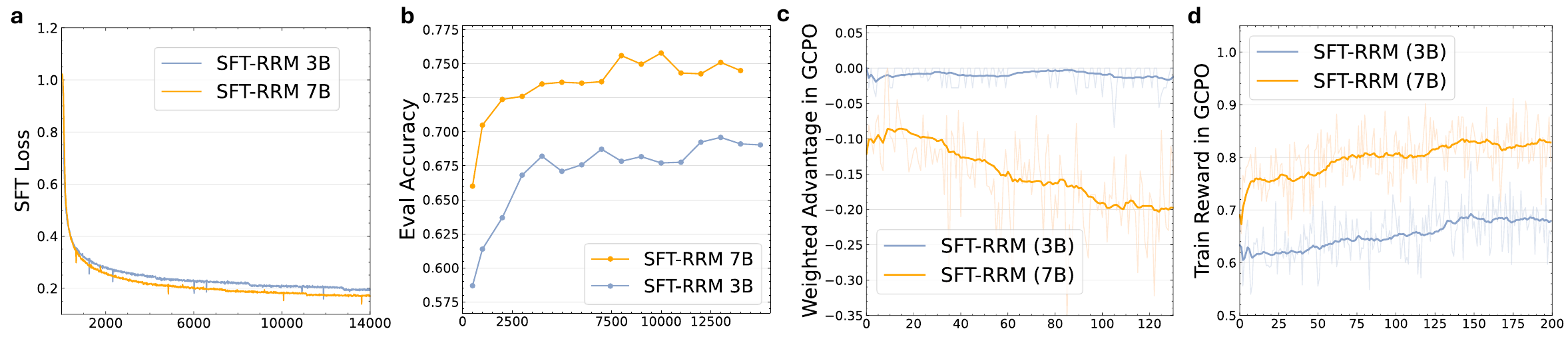} 
    \vspace{-1em}
    \caption{
        Training dynamics of RRMs. 
        \textbf{a}, SFT Loss, showing model convergence and scalability. 
        \textbf{b}, SFT evaluation accuracy for the RRMs, showing steady improvement. 
        \textbf{c}, Weighted advantage during GCPO training. The weighted advantage is defined as $\frac{1}{G}\sum_{i=1}^{G}\frac{A_{i}}{L_{i}}$, with $L$ represents the length of reasoning tokens. The negative value indicates it learns to generate longer reasoning traces for correct judgments. 
        \textbf{d}, Training reward during the GCPO phase, showing stable improvement and scalability.
    }
    \label{fig:rm_dynamics}
\end{figure*}

\subsubsection{Reasoning-Reinforced Reward Learning}
\label{sec:r3l}

Although the reward model possesses effective Chain-of-Thought (CoT) reasoning capabilities following the cold-start SFT phase, we observe that its judgments can be fallible. The model may exhibit hallucinations or struggle to accurately assess the magnitude of edits, such as incorrectly verifying a principle "move to the left of the figure" as successful, but the object has only slightly moved, as detailed in the Appendix. 

To further align the model with human preferences, we introduce a reinforcement learning phase.
A key challenge is that the RRM first generates a reasoning trace before producing a final score, making it difficult to optimize with standard scalar-reward formulations directly.



As illustrated in Fig.~\ref{fig:mainfig_v2}, this phase employs an inter-reward, intra-advantage based GRPO algorithm, which we term Group Contrastive Preference Optimization (GCPO), to refine the reward model using human-annotated preference pairs. In this framework, the reward model $R_\phi$ itself serves as the policy being optimized, where $\phi$ are its parameters. The "actions" consist of the generated reasoning trace and the final score, which are produced conditioned on the input quadruple data.

\vspace{-.5em}
\paragraph{Preference Data.}
For this phase, we construct a preference dataset, $\mathcal{D}$, through human annotation. The annotation process was as follows: annotators were presented with a source image $x_{\mathrm{ref}}$, an editing instruction $q$, and a pair of edited images. They were asked to choose which image was better, or to label them as "same" if they were of comparable quality or if a clear preference could not be established. The primary criteria for judgment were instruction fidelity and overall image quality. This process yielded a dataset of approximately 10,000 preference pairs $(x^w, x^l)$ for each context $c=(x_{\mathrm{ref}}, q)$, where $x^w$ denotes the preferred (winner) image and $x^l$ the non-preferred (loser) one. The "same" pairs were excluded from the GCPO training.

\paragraph{Win/Loss Ratio Rewards.}
We employ pairwise \emph{win/loss ratio} rewards derived from cross-group preference comparisons. 
For each preference pair $(x^w, x^l)$, the reward model $\mathbb{R}_\phi$ stochastically generates $N$ distinct reasoning traces and their corresponding scores $\{\tau_j^w\}_{j=1}^{N}$ and $\{\tau^l_j\}_{j=1}^{N}$ for each image:
\begin{align}
 \tau_j^w = \Phi\!\left(\mathbb{R}_\phi(x^w_j,c,\mathcal{P})\right), \quad \tau^l_j = \Phi\!\left(\mathbb{R}_\phi(x^l_j,c,\mathcal{P})\right),
\end{align}
where $\Phi(\cdot)$  is an operator that extracts the scalar score from the text output of $\mathbb{R}_\phi(\cdot,\cdot,\cdot)$ via rule-based parsing.

The per-sample win/loss ratios are then defined based on exhaustive pairwise comparisons between the two sets of scores, ignoring ties. The \textbf{win ratio} for a preferred candidate $\tau_g^j$ is the fraction of non-preferred candidates it scores \emph{higher} than. Symmetrically, the \textbf{loss ratio} for a non-preferred candidate $\tau_b^j$ is the fraction of preferred candidates that score \emph{lower} than it:
\begin{align}
r^w_j &= \frac{1}{N}\sum_{k=1}^{N} \Ind{\tau^w_j > \tau^l_k},\quad r^l_j = \frac{1}{N}\sum_{k=1}^{N} \Ind{\tau^l_j < \tau^w_k},
\end{align}
where $N$ denotes the number of reasoning traces generated, $\Ind{\cdot}$ is the indicator function.

\paragraph{Optimization with GCPO.}
After computing the win/loss ratio rewards $\{r^w_j\}_{j=1}^{N}$ and $\{r^l_j\}_{j=1}^{N}$ from cross-group comparisons, the original pairing between samples is disregarded for the optimization step. Instead, advantages are computed independently \emph{within} each rollout group (preferred or non-preferred). Although the rewards originate from paired comparisons, the loss is calculated by partitioning the rollouts into two distinct sets. The advantages are computed as follows:
\begin{align}
& \bar{r}^w = \frac{1}{N}\sum_{j=1}^{N} r^w_j, \qquad
\bar{r}^l = \frac{1}{N}\sum_{j=1}^{N} r^l_j, 
\qquad \nonumber \\
& A^w_j = r^w_j - \bar{r}^w, 
\qquad
A^l_j = r^l_j - \bar{r}^l.
\end{align}
Let $r_{t,j}^{w}(\phi)$ and $r_{t,j}^{l}(\phi)$ denote the per-token likelihood ratios for the $j$-th rollout and $t$-th token in the preferred and non-preferred groups, respectively. The objective function is the sum of the two groups' clipped surrogate losses, omitting the KL divergence term:
\begin{figure*}[t]
\vspace{-1.5em} 
\begin{equation}
\label{eq:gcpo_final}
\begin{split}
    \mathcal{L}_{\text{GCPO}}(\phi) = \mathbb{E}_{\dots\sim\mathcal{D}} \Bigg[ \frac{1}{2N} \sum_{j=1}^{N} \frac{1}{T} \sum_{t=0}^{T-1} \bigg( 
    &\min\Big( r_{t, j}^{w}(\phi)A^w_j, \text{clip}\big(r_{t, j}^{w}(\phi), 1-\epsilon, 1+\epsilon\big)A^w_j \Big) \\
    &+ \min\Big( r_{t,j}^{l}(\phi)A^l_j, \text{clip}\big(r_{t,j}^{l}(\phi), 1-\epsilon, 1+\epsilon\big)A^l_j \Big)
    \bigg) \Bigg].
\end{split}
\end{equation}
\vspace{-1.5em} 
\end{figure*}

\subsection{Reinforcement Learning for Image Editing}
\label{sec:rl_for_editing}
We employ the GRPO algorithm ~\citep{liu2025flow} and leverage our reasoning-reinforced reward model to provide fine-grained feedback to optimize the image editing model.
The editing model acts as the policy $\pi_\theta(\cdot,c)$ for each sampling step. Following the Group Relative Policy Optimization (GRPO) paradigm, optimizing is as follows:
For each conditioning context $c$ sampled from our dataset $\mathcal{D}$, the flow-based editing model $\pi_\theta(\cdot,c)$ generates a group of $G$ edited images $\{x_0^i\}_{i=1}^G$ along with their corresponding generation trajectories $\{(x_T^i, \dots, x_0^i)\}_{i=1}^G$, where $\{x_T^i\}_{i=1}^G$ is sampled from gaussian distribution.

Our verifiable reward model, $\mathbb{R}_\phi(\cdot,\cdot,\cdot)$ verifies and evaluates each generated image $x_0^i$ based on the context $c$ and the corresponding principle $\mathcal{P}$. 
Within each group, the advantage $A_i$ for the \textit{i}-th image is calculated by normalizing its reward against the mean and the standard deviation within a group:
\vspace{-0.5em}
\begin{equation}
A_i = \frac{\tau_i - \text{mean}\left(\{\tau_i\}_{i=1}^G\right)}{\text{std}\left(\{ \tau_i\}_{i=1}^G\right) + \epsilon_{\text{std}}},\quad \tau_i = \Phi\!\left(\mathbb{R}_\phi(x_0^i, c, \mathcal{P})\right),
\label{eq:advantage}
\end{equation}
where $\tau_i$ is the holistic reward score provided by our reasoning-based reward model for the \textit{i}-th sample, and $\epsilon_{\text{std}}$ is a small constant for numerical stability.
The GRPO training objective is to maximize the expected advantage, incorporating a clipped objective function to prevent excessively large policy updates and a KL-divergence penalty term to regularize the policy $\pi_\theta(\cdot,c)$ and keep it from deviating too far from a reference policy $\pi_{\text{ref}}(\cdot,c)$. 

This Edit-aware GRPO framework enables the editing model to directly optimize for human-perceived quality and instruction fidelity as captured by our verifiable reward model.
\vspace{1em}

\begin{figure*}[t!]
    \centering
    \includegraphics[width=0.96\textwidth]{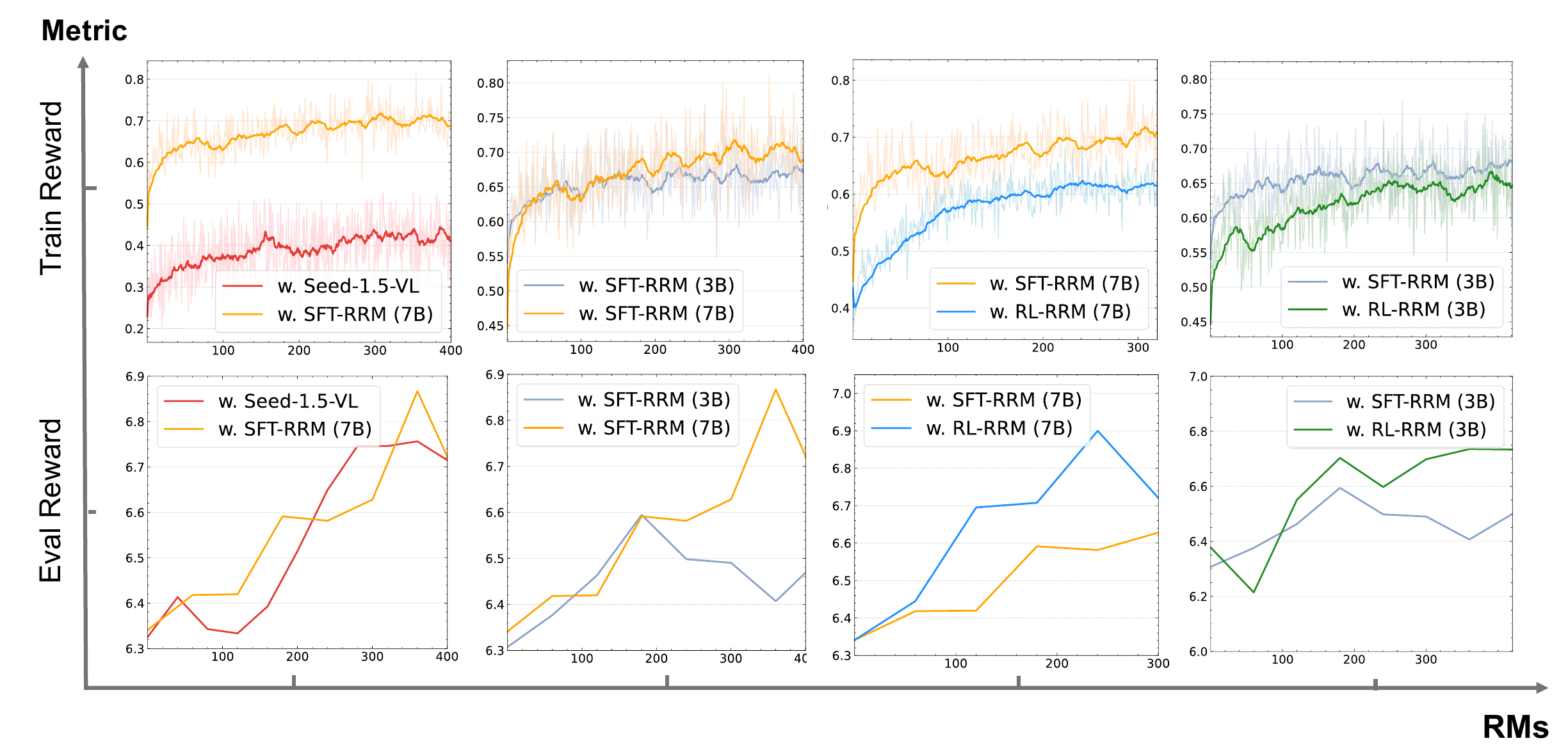} 
    \captionsetup{font=small}
    \vspace{-1.0em}
    \caption{
    Training dynamics of editing model optimization with different RRMs. The first row shows the training reward, and the second row shows the evaluation reward.
    Here, SFT-RRM denotes a reward model trained without GCPO, while RL-RRM denotes its counterpart trained with GCPO.  First column: our SFT-RRM (7B) produces a reward signal that is as stable and effective as the Seed-1.5-VL. Second column: the SFT-RRM 7B exhibits stronger scalability, providing more reliable supervision and yielding better performance than the SFT-RRM 3B. Third and fourth columns: refining the RRM with GCPO results in consistently higher evaluation rewards, indicating that the RRM trained with GCPO acts as a stricter and more robust evaluator.
    }
    \label{fig:edit_dynamics}
\end{figure*}

\section{Experiments}
\subsection{Experimental setups}

\paragraph{Benchmark and Metrics.}
For reward model evaluation, we curated a high-quality and diverse set of 5,000 reference images and instructions from the same public image-editing benchmark. 
We then utilized various models, including SeedEdit-3.0~\citep{wang2025seededit}, BAGEL~\citep{deng2025emerging}, and FLUX.Kontext~\citep{batifol2025flux}, to produce several edited outputs for each input. Finally, these generated outputs were manually annotated using pairwise preference comparisons. 
Annotators were also allowed to mark a pair as ``same'' when the two edits were of comparable quality or when no reliable preference could be established. These ambiguous pairs were excluded when evaluating verifier accuracy, yielding a cleaner pairwise benchmark.
The accuracy of the reward model in predicting these human-annotated preferences is used as our evaluation metric. We also report performance on the public EditRewardBench~\citep{luo2025editscore} for a comprehensive comparison.
For image editing model evaluation, we adopt GEdit-Bench-EN~\citep{liu2025step1x}, a standardized benchmark with multi-dimensional automatic metrics. Following the original protocol, we report scores across three key aspects, each assessed by GPT-4.1: semantic consistency (SC), which measures how well the edited image aligns with the given instruction; perceptual quality (PQ), which captures the visual fidelity of the edited image; and overall score (O), computed as the geometric mean of SC and PQ. In addition, we report SC scores for different categories, as presented in Tab.~\ref{tab:model_evaluation}.


\paragraph{Implementation Details.}
Our RRM is built on the open-source Qwen-VL-2.5~\citep{bai2025qwen2}.
In the cold-start SFT phase in Sec.~\ref{sec:cold_start_reward}, we constructed editing pairs using a mixture of models, including SeedEdit3.0~\citep{wang2025seededit}, BAGEL~\citep{deng2025emerging}, and FLUX.Kontext~\citep{batifol2025flux}.
For the \gcpo{} phase in Sec.~\ref{sec:r3l}, we further collected 10k human-annotated preference pairs, which is less than 1\% of the SFT-scale training data. Therefore, the gains from GCPO are mainly attributable to better human alignment rather than increased data volume.
For editing model optimization, we apply our \editrone{} framework to two strong open-source models: FLUX.Kontext~\citep{batifol2025flux} and Qwen-Image-Edit~\citep{wu2025qwen}.
The models are optimized using the \grpo{} strategy described in Sec.~\ref{sec:rl_for_editing}, with our trained \rrm{} serving as the reward signal.
We adopt Flow-GRPO~\citep{liu2025flow} with a group size of $G=24$ and a KL penalty coefficient of $\beta=0.04$.
Although GCPO requires rollouts from both preferred and non-preferred groups, the overall training cost remains manageable in practice due to the small rollout group size and efficient packed inference.
\subsection{Reward Model Performance}
Our proposed \editrone{} framework yields a state-of-the-art reward model for predicting human preferences. As shown in Tab.~\ref{tab:full_rm_results}, our 7B Edit-RRM, trained via our full two-stage pipeline, achieves a top accuracy of \textbf{82.2\%}. 
This result significantly surpasses strong closed-source APIs like Seed-1.5-VL (79.3\%) and demonstrates the effectiveness of our training strategy.
\begin{table}[h] 
    \centering
    \caption{Accuracy on our internal benchmark. \textbf{T}, \textbf{V}, and \textbf{T+V} denote Think, Verify, and Think+Verify, respectively.}
    \label{tab:full_rm_results}
    \renewcommand{\arraystretch}{1.2}
    \setlength{\tabcolsep}{3pt} 
    \small
    \begin{tabular*}{\columnwidth}{@{\extracolsep{\fill}}l ccc c}
        \toprule
        \textbf{Model} & \textbf{T} & \textbf{V} & \textbf{T+V} & \textbf{GCPO} \\
        \midrule
        \multicolumn{5}{l}{\textit{Inference w. API Baselines}} \\
        Seed-1.5-VL~\citep{guo2025seed1}  & 72.2\% & --- & 79.3\% & --- \\
        Seed-1.6-VL~\citep{guo2025seed1}  & 71.2\% & 69.4\% & 77.2\% & --- \\
        \midrule
        \multicolumn{5}{l}{\textit{Our Method (SFT \& RL Stages)}} \\
        Qwen-7B (VIESCORE) & \multicolumn{3}{c}{68.3\%} & --- \\
        \midrule
        Qwen-3B  & 64.1\% & 66.1\% & 69.3\% & 72.0\% \\
        Qwen-7B  & 68.9\% & 70.9\% & 75.4\% & \textbf{82.2\%} \\
        \bottomrule
    \end{tabular*}
\end{table}
\paragraph{Verifier vs. Scorer on Public Benchmarks.}
On the public EditRewardBench, our verifier-based RRM consistently outperforms the holistic scorer baseline. As shown in Tab.~\ref{tab:editscore_benchmark_comparison}, our SFT-RRM already surpasses EditScore-7B (73.3\% vs. 65.9\%), and GCPO further improves the final RL-RRM to 78.2\%. Since EditRewardBench is independently constructed from our internal pipeline, this gain indicates that our improvement is not due to internal benchmark bias. 
\begin{table*}[t]
\centering
\caption{
Detailed performance comparison on the GEdit-Bench-EN (Full set). Higher scores are better. \textbf{Bold} scores highlight the best result within each model family. 
Columns 1–11 report SC scores for different editing categories (see Appendix for details).
}
\label{tab:model_evaluation}
\renewcommand{\arraystretch}{1.25} 
\resizebox{\textwidth}{!}{
\begin{tabular}{l *{11}{S[table-format=1.2]} S[table-format=1.2] S[table-format=1.2] S[table-format=1.2]}
\toprule
\multirow{2}{*}{\textbf{Model}} & \multicolumn{11}{c}{\textbf{Category SC}} & \multicolumn{3}{c}{\textbf{Overall}} \\
\cmidrule(lr){2-12} \cmidrule(lr){13-15}
& {1} & {2} & {3} & {4} & {5} & {6} & {7} & {8} & {9} & {10} & {11} & {SC $\uparrow$} & {PQ $\uparrow$} & {O $\uparrow$} \\
\midrule
\multicolumn{15}{l}{\textbf{Edited Models}} \\
Step-Edit~\citep{liu2025step1x}       & 8.77 & 8.90 & 7.52 & 4.35 & 4.10 & 7.73 & 8.56 & 7.81 & 8.26 & 2.82 & 7.30 & 6.53 & 6.72 & 5.90 \\
UniPic2~\citep{wei2025skywork}        & 8.07 & 8.70 & 6.75 & 3.57 & 4.78 & 7.13 & 8.36 & 8.36 & 7.87 & 5.39 & 7.97 & 6.84 & 7.24 & 6.41 \\
Bagel~\citep{deng2025emerging}        & 8.54 & 8.32 & 7.42 & 4.97 & 5.07 & 7.71 & 8.75 & 8.03 & 8.22 & 7.14 & 6.62 & 7.32 & 7.02 & 6.65 \\
GPT-4o                               &      &      &      &      &      &      &      &      &      &      &      & 7.74 & 8.13 & 7.49 \\
\midrule
\multicolumn{15}{l}{\textbf{FLUX.Kontext Family}~\citep{batifol2025flux}} \\
FLUX.Kontext  & 7.65 & 8.17 & 6.90 & 3.02 & 3.54 & 6.86 & 7.43 & 7.54 & 6.95 & 5.14 & 7.85 & 6.27 & 7.25 & 5.77 \\
\rowcolor{gray!10}
RL w. SeedVLM-1.5                    & 8.23 & 8.41 & 7.00 & \textbf{5.00} & 4.05 & 7.17 & 7.93 & 7.60 & 7.11 & 5.51 & 8.60 & 6.74 & 6.44 & 6.03 \\
\rowcolor{gray!10}
RL w. SFT-RRM (3B)                   & 7.48 & 8.25 & 6.65 & 3.20 & 4.01 & 7.25 & 7.77 & 7.15 & 7.61 & 5.70 & 7.90 & 6.52 & 6.26 & 5.63 \\
\rowcolor{gray!20}
RL w. RL-RRM (3B)                    & 8.13 & 7.65 & \textbf{7.33} & 3.63 & 3.70 & 7.25 & 7.78 & 7.65 & \textbf{7.68} & 6.03 & \textbf{7.93} & 6.67 & 7.09 & 6.10 \\
\rowcolor{gray!10}
RL w. SFT-RRM (7B)                   & \textbf{8.25} & 8.52 & 7.10 & 3.85 & 3.68 & 7.27 & \textbf{8.26} & 7.68 & 7.38 & 6.44 & 7.80 & 6.81 & \textbf{7.25} & 6.20 \\
\rowcolor{gray!20}
RL w. RL-RRM (7B)                    & 8.15 & \textbf{8.62} & 6.82 & 4.42 & \textbf{4.22} & \textbf{7.30} & 8.21 & \textbf{7.99} & 7.41 & \textbf{6.44}  & 7.82 & \textbf{6.86} & 7.20 & \textbf{6.24} \\
\midrule
\multicolumn{15}{l}{\textbf{Qwen-Edit Family}~\cite{wu2025qwen}} \\
Qwen-Edit                            & \textbf{8.85} & 9.02 & \textbf{8.20} & 4.01 & 6.04 & 7.61 & \textbf{8.80} & 8.32 & 8.74 & \textbf{9.00} & \textbf{8.37} & 7.94 & \textbf{7.78} & 7.45 \\
\rowcolor{gray!20}
RL w. RL-RRM (7B)                & 8.75 & \textbf{9.05} & 8.10 & \textbf{4.62} & \textbf{6.17} & \textbf{7.76} & 8.72 & \textbf{8.43} & \textbf{8.79} & 8.69 & 8.32 & \textbf{7.99} & 7.76 & \textbf{7.50} \\
\bottomrule
\end{tabular}
}
\end{table*}
\paragraph{Principled Data Curation and SFT.}
The foundation of our model's performance is a carefully curated SFT dataset. Tab.~\ref{tab:full_rm_results} shows that the full SFT pipeline with both ``Think'' and ``Verify'' (T+V) consistently performs best before GCPO. For Qwen-7B, accuracy improves from 68.9\% with ``Think'' only to 75.4\% with ``Think+Verify'', highlighting the importance of principled decomposition and rigorous filtering. As a reference baseline, SFT data generated with VIESCORE prompts~\cite{ku2024viescore} achieves 68.3\%. These results support our claim that each component in the cold-start pipeline is important for reward-model supervision.
\paragraph{Ablation on GCPO and Benchmarks.} 
We further analyze the impact of \gcpo{} on the public EditScore-RM benchmark (Tab.~\ref{tab:editscore_benchmark_comparison}). GCPO provides a clear performance boost, improving accuracy from 73.3\% to 78.2\%, which shows that it is critical for refining the model's alignment with human preferences. Notably, our final RRM also outperforms EditScore-7B with inference scaling. Since the GCPO stage uses only 10k human preference pairs---less than 1\% of the SFT-scale data, whose gain is mainly attributable to better human alignment.
\paragraph{Impact of Principled Data Curation.}
Our initial experiments (not shown in the main tables for brevity) confirmed that both the ``Think" (reasoning generation) and ``Verify" (external filtering) components are critical. Removing the ``Verify" step led to a significant accuracy degradation, highlighting the importance of rigorous data filtering. Meanwhile, the reasoning traces from the ``Think" step provide essential supervisory signals that enhance the model's evaluative capabilities.
\subsection{Image Editing Performance}
Fig.~\ref{fig:edit_dynamics} shows that our RRMs provide stable reward signals, while GCPO-refined RRMs yield higher evaluation rewards and act as stricter evaluators.
\vspace{-.5em}
\paragraph{Overall Performance.}
As shown in Tab.~\ref{tab:model_evaluation}, our framework demonstrates strong performance on both model families. Optimizing FLUX.Kontext with our RL-RRM (7B) boosts its Overall Score (O) from 5.77 to 6.24 and its Semantic Consistency (SC) score from 6.27 to 6.86. This result exceeds others, confirming that our RRM is a highly effective and competitive reward source for policy optimization.
Meanwhile, we carried out experiments on the SOTA open resource, Qwen-Edit, whose overall score showed a modest improvement from 7.45 to 7.50. This is largely because the baseline model already benefits from Best-of-N scaling~\cite{luo2025editscore}. However, the validity of Edit-R1 is highlighted in its ability to address the model's specific weaknesses.
Notably, as shown in the final row of Tab.~\ref{tab:model_evaluation}, our framework yields a significant 15.2\% relative gain (from 4.01 to 4.62) in the challenging Motion Change category, demonstrating its effectiveness in enhancing performance even on highly optimized models. These gains are further supported by human evaluation: FLUX.Kontext optimized with our RL-RRM (7B) achieves a GSB score of +23.2 against the original FLUX.Kontext baseline. The user-study protocol and full human-evaluation details are provided in Appendix~\ref{sec:human_eval}, and additional qualitative examples are shown in Fig.~\ref{fig:qualitative_results}.
\paragraph{Impact of GCPO and Scalability.} 
As shown in Tab.~\ref{tab:model_evaluation}, RRM with GCPO consistently outperforms its SFT counterpart. Training curves in Fig.~\ref{fig:edit_dynamics} further reveal the mechanism. Although RL-RRMs provide lower training rewards, they yield higher evaluation rewards, indicating that GCPO transforms the reward model into a stricter and more reliable evaluator. This stricter supervision pushes editing models to adhere more closely to human preferences and achieve higher-quality outputs.
\begin{table}[t!]
    \centering
    \caption{Comparison of our RRM against the baseline on the EditReward benchmark. All results are for 7B models.}
    \label{tab:editscore_benchmark_comparison}
    \renewcommand{\arraystretch}{1.15}
    \small
    \begin{tabular}{lc}
        \toprule
        \textbf{Method} & \textbf{Accuracy (\%)} \\
        \midrule
        \multicolumn{2}{l}{\textit{Baseline Model}} \\
        EditScore-7B & 65.9\% \\
        \quad + \textit{inference scaling} & 72.7\% \\
        \midrule
        \multicolumn{2}{l}{\textit{Our Method}} \\
        Our RRM (SFT only) & 73.3\% \\
        Our RRM (SFT + GCPO) & \textbf{78.2\%} \\
        \bottomrule
    \end{tabular}
\end{table}
\paragraph{Qualitative Analysis.} 
Beyond quantitative metrics, the qualitative improvements are equally compelling. As shown in the Appendix (Fig.~\ref{fig:qualitative_results}), models optimized with our framework exhibit markedly better instruction adherence and visual fidelity. 
In challenging edits such as subject addition/removal and motion change, our method successfully follows the instruction, whereas the baseline fails. For localized edits such as Color Alter, our approach precisely modifies the target object without introducing global color shifts or artifacts. These qualitative results provide concrete evidence of the practical effectiveness of our framework.

\section{Conclusion}

We introduce Edit-R1, a novel framework designed to enhance image editing through Reinforcement Learning from Human Feedback. Its core is a Verifier-based Reasoning-Enhanced Reward Model (RRM), trained via a ``cold-start" SFT phase and our innovative GCPO algorithm, which achieves evaluation accuracy that surpasses powerful proprietary models. By integrating this powerful RRM with a GRPO-based RL algorithm, our framework substantially enhances the instruction-following capabilities of state-of-the-art editing models. 



\section*{Acknowledgements}
This work was supported by Hong Kong Research Grants Council under the NSFC/RGC Collaborative Research Scheme (Grant CRS\_HKU703/24).

\clearpage
\bibliographystyle{plainnat}
\bibliography{main}

\clearpage
\appendix
\section{System prompt}
\subsection{System Prompt for Decomposing Principles}
\label{sec:appendix_prompt}
In practice, the prompt is used in an in-context learning manner with expert-written decomposition examples. We maintain a pool of 60 expert-authored exemplars and randomly sample 4 of them for each query to improve diversity and robustness in principle generation.

To automatically decompose image editing principles, we designed a detailed system prompt for a large vision-language model. This prompt utilizes a few-shot learning approach, providing the model with a complete example before presenting it with a new task. The structure is designed to define the model's role, specify task requirements and output format, and provide contextual examples. The prompt used for this purpose is detailed below.
\begin{lstlisting}[style=promptstyle, caption={The detailed system prompt for decomposing principles given the source image and edit instruction.}, label=lst:prompt_en_detailed]
You are an expert image editing evaluator. Your task is to generate evaluation points for a new image editing task.
### Reference Example:
Example: Instruction: Convert the original image to anime style
Principles:
[
  {
    "question": "Is the generated image converted to an anime style based on the original image?",
    "category": "Instruction Following"
  },
  {
    "question": "Does the character in the generated image retain the hair and facial features from the original image?",
    "category": "Feature Preservation"
  },
  {
    "question": "Does the character's clothing in the generated image retain the features from the original image?",
    "category": "Feature Preservation"
  },
  {
    "question": "Does the character's pose in the generated image remain consistent with the original image?",
    "category": "Feature Preservation"
  },
  {
    "question": "Do the background elements like the table, sofa, bed, and window retain their original features and layout?",
    "category": "Feature Preservation"
  },
  {
    "question": "Apart from the main background elements mentioned, are other details from the original image preserved?",
    "category": "Feature Preservation"
  },
  {
    "question": "Is the generated image free of significant structural problems?",
    "category": "Image Quality"
  },
  {
    "question": "Is the clarity and overall quality of the generated image good?",
    "category": "Image Quality"
  },
  {
    "question": "Does the scene with the character in the generated image look natural?",
    "category": "Image Quality"
  }
]
### Task Requirements:
Generate 10 evaluation points for the new image editing task, with the following distribution:
1. 3-4 points for "Instruction Following" (to assess the implementation of the edit).
2. 3-4 points for "Feature Preservation" (to assess the retention of original features).
3. 2-3 points for "Image Quality" (to assess the quality of the resulting image).
### Output Format:
A JSON array, where each element contains a 'question' field and a 'category' field.
### New Task:
Instruction: {Edit Instruction}
Image: <image>
Please generate all evaluation points:
\end{lstlisting}

\subsection{System Prompt for Reward Model Evaluation}
\label{sec:appendix_prompt_rrm}
To quantitatively score the edited image, we employ a Verifier-based Reasoning Reward Model (RRM). The RRM is guided by a detailed system prompt designed to act as a professional evaluator. This prompt instructs the model to firstly assess the edited image based on the decomposed principles, considering the edit instruction, and performing a holistic analysis of the output quality. The prompt defines a structured evaluation process, including rule definitions, an execution flow, and a strict output format. The complete system prompt provided to the RRM is detailed below.
\begin{lstlisting}[style=promptstyle, caption={The prompt used for the Reward Model (RRM) to evaluate the quality of edited images. This prompt guides the model to score based on predefined decomposed principles.}, label=lst:prompt_rrm_en]
You are a professional evaluation point analyst and image editing evaluator. Your task is to analyze whether a generated image meets a given set of evaluation points, based on the input image and an edit instruction. You must also use divergent thinking based on these points to holistically evaluate the model's editing performance. Your evaluation should not be based solely on the magnitude of the edit; instead, you must conduct a comprehensive, side-by-side comparison for each evaluation point. If an evaluation point is not met, you must assess the difficulty and complexity of revising the edited image to meet it. Furthermore, you must consider whether elements not mentioned in the instruction or evaluation points (such as the background or secondary subjects) have undergone unreasonable changes. If they were not supposed to change but did, points should be deducted accordingly.
## Input:
- Original Image: <image>
- Edited Image: <image>
- Edit Instruction: {{EDIT_INSTRUCTION}}
- Evaluation Points: {{EXAM_POINTS}}
## Rule Definition:
- For each evaluation point (e.g., "Was the scene changed from indoors to outdoors?"), you can only assign a score of 0 (not met) or 1 (met). For edits involving a range (e.g., far to near, left to right, male to female, fat to thin), a significant change is required to be considered 'met' unless the magnitude is specified. When considering relative positions, if an object faces the camera, the object's left is considered the right side from the viewer's perspective.
- If you are uncertain about an evaluation point, score it as 0 (not met) and incorporate this uncertainty into your subsequent reasoning for the final score.
- The final score should not be solely dependent on the average of the evaluation point scores. The final score can be any value from 0 to 10, not just integers like 0, 5, 8, or 10. If you are not confident about an integer score, use a decimal. If an evaluation point contradicts the edit instruction (e.g., preserving a watch while the instruction is to lower the hand, which would hide it), this point should be ignored when calculating the final score. The consistency of newly revealed areas due to object movement requires special attention, while focusing on the consistency of originally un-occluded parts.
- A perfect score on the evaluation points does not guarantee a perfect final score. You must assess if the edited image is directly usable, if the edit magnitude is appropriate, and if it meets psychological expectations. Also, consider if unmentioned elements have changed unreasonably.
- Crucially, if the edited image is nearly identical to the original (i.e., no edit was performed), assign a score of 0. If the instruction involves a single edit, that edit is the most critical part of the task; if the similarity is too high, the image requires major correction, so score it 0. If the edited image has white borders, score it 0.
- Preserve class information. For example, consistency should be judged based on 3D integrity of material and structure. Even if the viewing angle changes, if it's the same object, consistency is considered good. Prioritize the consistency of the main subject, then secondary subjects/objects. Penalize minor inconsistencies but not heavily if the main subject's consistency is maintained. However, for removal tasks, the object must be completely removed, so pay close attention to positional information of small objects or subjects.
- When dealing with positional information, you must output bounding box coordinates in your thought process.
- For positional changes (e.g., from left to right), a significant shift is required; a minor move is not sufficient.
- When evaluating human pose, strictly determine left and right based on the person's orientation.
- If an edit instruction has N points and one is not met, the deduction should be based on the cost of re-editing the current image to fix that specific point. Deduct more points for fixes that require more information or have a lower probability of success.
- When determining the final score, consider the completion status of multiple key points in the instruction, with a focus on the core directive. For any unmet point, think about the future editing cost (e.g., needing more conditions, more information, or modifying more pixels). Compare this cost hypothetically with the cost of completing other unmet conditions to judge the deduction.
- When an evaluation point contradicts the edit instruction (e.g., requiring consistent color tone during a style transfer, or preserving details on a limb that is moved out of view), prioritize achieving the edit instruction.
- Also, when thinking about the final score, consider unmentioned aspects like the main subject, secondary subjects, and background. If they were not supposed to change but did, or if they changed but are inconsistent, this is a hallucination and should be penalized more heavily.
- Additionally, check for quality issues. The image should not have white borders (minor deduction). Also, check for over-sharpening, oversaturation, or color cast.
- When reasoning about the final score, re-check the following aspects in order of importance: quantity, action/state, negations/comparatives, composition/form/function, material, position/state (e.g., hanging), composition, main subject, environment.
## Execution Flow:
Please follow these steps strictly and sequentially. Do not skip or omit any step:
1. For each evaluation point provided in the format `[{'question':, 'category': }]`, evaluate and score it based on a comparison of the before/after images and the edit instruction, strictly adhering to the scoring standards in the [Rule Definition].
2. Based on the above, assign a final score to the edited image from 0 to 10. 0 means completely unusable (e.g., severe artifacts, very difficult to fix manually). 5 means partially usable (some good aspects but far from ready). 8 means nearly usable (minor artifacts, inconsistencies, or instruction deviations that can be fixed with minor manual intervention).
3. When positional changes are involved, output bounding box coordinates in your thought process to reflect your analysis of the position, and then judge if the edit is valid based on the scale of change defined in the rules.
4. Finally, assess the difference between the before and after images to confirm that an edit has actually occurred.
## Output Format:
Produce the output in the following sequence: scores for each evaluation point, the average score of the evaluation points, and finally, the reasoned final score for the generated image.
`[{'question':, 'score': }, ...], {"average_score": } <score> <\score>`
\end{lstlisting}

\subsection{System Prompt for VLM Verification}
\label{sec:verification}
Our data annotation pipeline incorporates a VLM-based verification stage to generate high-quality, fine-grained evaluation data. This process is divided into two steps, each guided by a specific system prompt: **Verification** and **Selection**.
First, a powerful VLM acts as a **Verifier**. It is presented with the source image, the edited image, the instruction, and a list of evaluation points. Crucially, it also receives "reference intermediate judgments"---a collection of Chain-of-Thought (CoT) reasoning excerpts and per-point predictions from multiple candidate models. The Verifier's task is to critically and objectively assess these materials to produce a "gold standard" 0/1 judgment for each evaluation point, effectively acting as an expert human annotator.
Second, another VLM acts as a **Selector**. It receives the newly created gold standard and the raw predictions from all candidate models. Following a strict, deterministic ruleset, it calculates the accuracy for each candidate and selects the best-performing one. This two-step process ensures both the quality of the annotations and the objective selection of the most accurate model output.
The prompts for both the Verifier and the Selector are detailed below.
\begin{lstlisting}[style=promptstyle, caption={System prompt for the VLM Verifier. It instructs the model to act as a strict inspector to generate gold-standard annotations.}, label=lst:prompt_vlm_verifier]
You are a strict image editing verification inspector. Your input includes: an original image, an edited image, an edit instruction, a list of evaluation points, and "reference intermediate judgments" (which are per-point predictions and brief reasoning summaries from multiple candidate models).
Your task is to objectively verify the edits based solely on the images and text, providing a gold-standard judgment (0 or 1) for whether each evaluation point was met, along with a one-sentence reason.
Note: The reference intermediate judgments are for reference only and must not be copied. If the references contradict the images and text, the images and text are the ground truth.
[Rules]
- Each evaluation point can only be scored 0 (not met) or 1 (met).
- If the required magnitude of a change is not specified, a "significant change" is required to be considered 'met'. (e.g., positional changes of less than 10% of the image dimensions are considered insufficient).
- For position-related points, mention the approximate region or bounding box in the reason. A person's left and right are determined by their facing direction.
- If the original and edited images are nearly identical => evaluation points related to the core edit instruction are judged as 0.
- Issues like white borders, severe sharpening, oversaturation, color cast, or structural artifacts can be considered, but the primary task is the per-point 0/1 judgment.
- 'Remove' tasks require complete removal. Prioritize the consistency of the main subject before considering smaller objects.
- If an evaluation point contradicts the edit instruction, prioritize the edit instruction.
## Input:
- Original Image: <image>
- Edited Image: <image>
- Edit Instruction: {{EDIT_INSTRUCTION}}
- Question Points: {{QUESTION_POINTS}}
- CoT-1: {{CoT}}
...
- CoT-N: {{CoT}}
[Output]
- Output ONLY a single JSON object (do not output any text outside the JSON), in the following format:
{
  "gold": [
    {"question": "Evaluation point text 1", "gold": 0 or 1, "reason": "Brief reason (can include approximate bbox)"},
    ...
  ]
}
The length and order of the 'gold' array must match the input list of evaluation points.
\end{lstlisting}

\section{Inference result of RRM}
\label{append:rrm}
\begin{figure*}[t]
    \centering
    \begin{subfigure}[b]{0.48\textwidth}
        \centering
        \includegraphics[width=\linewidth]{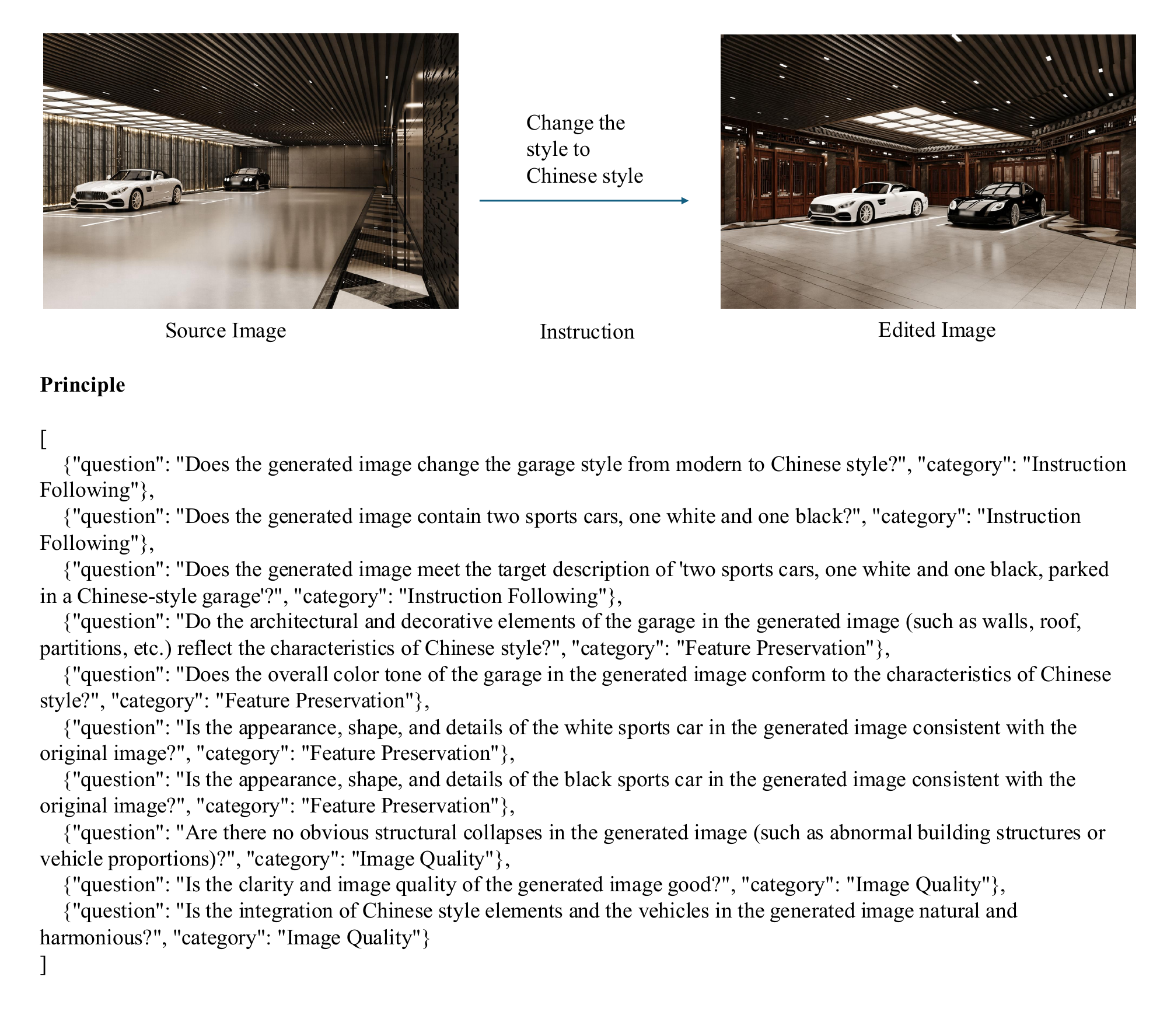}
        \caption{The input quadruple for the RRM, consisting of the source image, edit instruction, and the decomposed principles.}
        \label{fig:rrm_input_demo}
    \end{subfigure}
    \hfill 
    \begin{subfigure}[b]{0.48\textwidth}
        \centering
        \includegraphics[width=\linewidth]{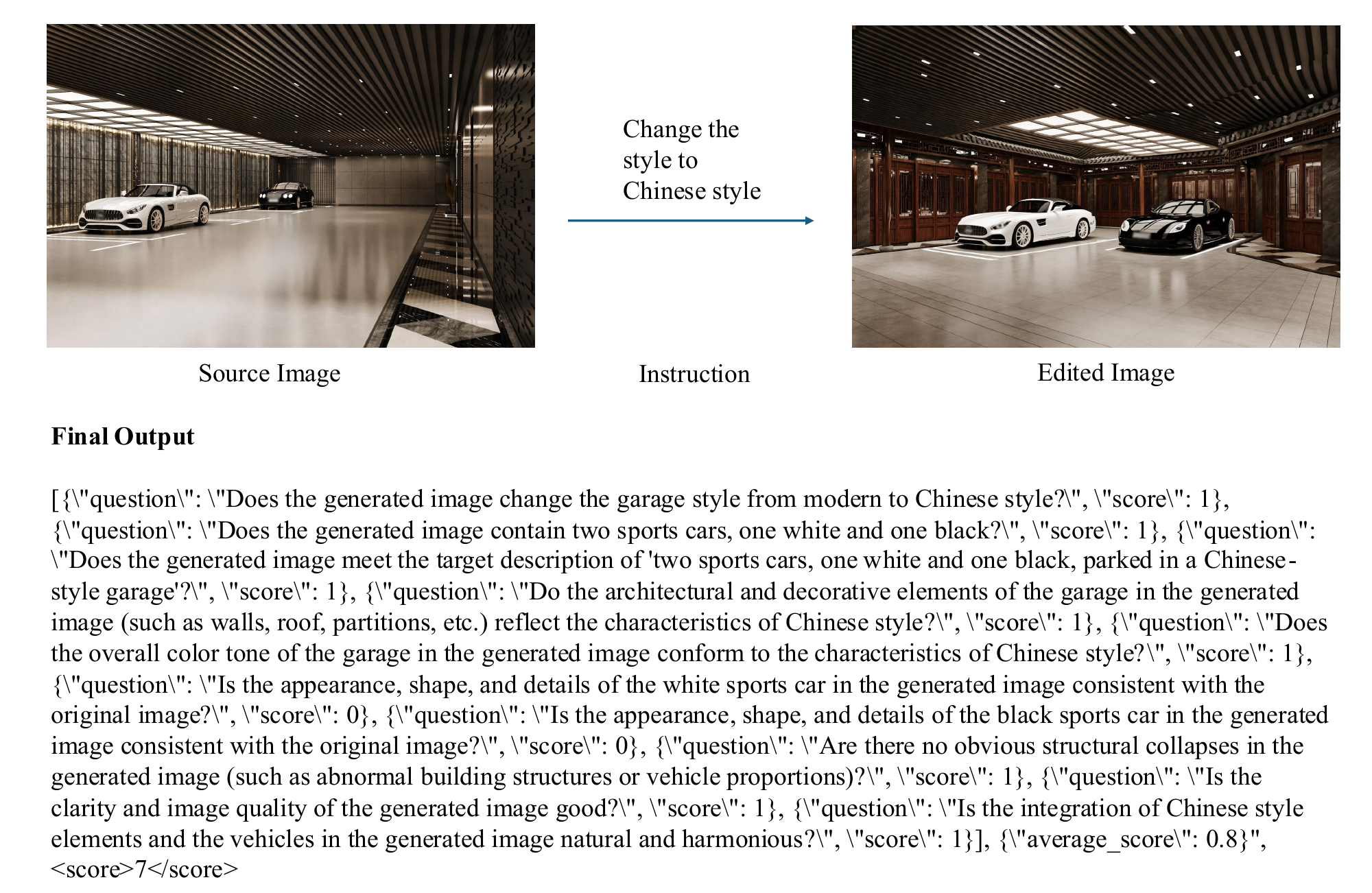}
        \caption{The final summarized response from the RRM after its reasoning process, providing the scores for each principle and a final score.}
        \label{fig:rrm_output_response}
    \end{subfigure}
    \caption{
        Illustration of the Verifier-based Reasoning Reward Model (RRM) inference process. (a) shows the input quadruple, which includes the source image, the edit instruction, and the decomposed principles for evaluation. (b) shows the final summary output from the RRM, containing the score for each principle and the final comprehensive score for the edited image.
    }
    \label{fig:rrm_inference_combined}
\end{figure*}

This section provides a concrete example of the inference process of our Verifier-based Reasoning Reward Model (RRM). The model receives a quadruple data point (source image, instruction, edited image, and principles) and generates a detailed chain-of-thought analysis to produce a final score. 
Fig.~\ref{fig:rrm_inference_combined} illustrates the input and final summarized output of this process. Following the figure, Listing~\ref{lst:reasoning_rrm} presents the complete, unabridged reasoning that leads to the final score.

\begin{lstlisting}[
    style=promptstyle, 
    caption={The detailed chain-of-thought reasoning from the RRM for a given task. This demonstrates how the model evaluates each principle to arrive at a final score.}, 
    label=lst:reasoning_rrm
]
The user now needs to analyze the score for each evaluation point. Let's look at them one by one:
First point: "Does the generated image change the garage style from modern to Chinese style?". Looking at the original image, it's a modern style (metal, clean lines). After editing, there are wooden doors and windows, and Chinese-style roof decorations (like tiles, wooden frames), so the style has changed. Score: 1.
Second point: "Does the generated image contain two sports cars, one white and one black?". The original image has a white and a black car, and the edited image does too. Score: 1.
Third point: "Does the generated image meet the target description of 'two sports cars, one white and one black, parked in a Chinese-style garage'?". The style is Chinese, and there are two cars, so it meets the description. Score: 1.
Fourth point: "Do the architectural and decorative elements of the garage in the generated image (such as walls, roof, partitions, etc.) reflect the characteristics of Chinese style?". The walls have wooden Chinese-style doors and windows, the roof has Chinese tiles and wooden grilles, and the partitions are also Chinese-style wood. So, it reflects the characteristics. Score: 1.
Fifth point: "Does the overall color tone of the garage in the generated image conform to the characteristics of Chinese style?". Chinese style often has warm, woody tones. The edited image has many wooden elements and a warmer tone, which is consistent. Score: 1.
Sixth point: "Is the appearance, shape, and details of the white sports car in the generated image consistent with the original image?". Looking at the white sports car, was the original a convertible? The edited one is a hardtop? The styling details have changed, such as the roof and body lines, so it is not consistent. Score: 0.
Seventh point: "Is the appearance, shape, and details of the black sports car in the generated image consistent with the original image?". The original black car was a different model. The edited black car's styling has changed significantly (front fascia, body lines), so it is not consistent. Score: 0.
Eighth point: "Are there no obvious structural collapses in the generated image (such as abnormal building structures or vehicle proportions)?". The building structure and vehicle proportions look normal, with no collapses. Score: 1.
Ninth point: "Is the clarity and image quality of the generated image good?". The image is clear and the quality is good. Score: 1.
Tenth point: "Is the integration of Chinese style elements and the vehicles in the generated image natural and harmonious?". The combination of Chinese architecture and modern sports cars does not look out of place visually; the integration is natural. Score: 1.
Now, calculate the score for each point:
1. Does the generated image change the garage style from modern to Chinese style?: 1
2. Does the generated image contain two sports cars, one white and one black?: 1
3. Does the generated image meet the target description of 'two sports cars, one white and one black, parked in a Chinese-style garage'?: 1
4. Do the architectural and decorative elements... reflect the characteristics of Chinese style?: 1
5. Does the overall color tone... conform to the characteristics of Chinese style?: 1
6. Is the appearance... of the white sports car... consistent with the original image?: 0
7. Is the appearance... of the black sports car... consistent with the original image?: 0
8. Are there no obvious structural collapses...?: 1
9. Is the clarity and image quality... good?: 1
10. Is the integration of Chinese style elements and the vehicles... natural and harmonious?: 1
Then calculate the average score: (1+1+1+1+1+0+0+1+1+1)/10 = 8/10 = 0.8
Next, consider the final score. The core instruction is to change to a Chinese style. The style change was successful (points 1-5, 8-10 are met), but the vehicle details (points 6, 7) were not preserved. However, the instruction is mainly about changing the style; the vehicles are to be present, not to have their details preserved. Therefore, the core instruction was highly accomplished, and the image quality is good. The points were deducted for vehicle details, but the instruction did not require the vehicle details to be consistent, only that two cars should be present. So the deduction for points 6 and 7 might be due to feature preservation, but the instruction was about style change. So these two points are for feature preservation. Perhaps the user wanted the vehicle appearance to remain unchanged? But the edit instruction is to change the style to Chinese, so is it necessary to preserve vehicle details? Looking at the instruction, it says "change the style to Chinese style," so whether the vehicle appearance is preserved is not the core of the instruction, but the evaluation points include feature preservation. So this needs to be considered.
The total score for all points is 8, and the average score is 0.8. Finally, considering the final score, it can be given a 7. Although the vehicle details have changed, the style change was successful, the image quality is good, and the integration is natural. It is close to usable, with minor issues (vehicle details) that are acceptable.
\end{lstlisting}
This detailed reasoning is then summarized by the RRM to produce the final scores shown in Fig.~\ref{fig:rrm_output_response}.

\section{Category label in quantitative results}
Category 1-11 represent Background Change, Color Alteration, Material Modification, Motion Change, Portrait Beautification, Style Transfer, Subject Addition, Subject Removal, Subject Replacement, Text Modification, Tone Transformation.

\section{Human Evaluation}
\label{sec:human_eval}

To validate that the automatic GPT-based metrics are aligned with human perception, we conducted a human study comparing FLUX.Kontext optimized by our RL-RRM (7B) against the original FLUX.Kontext baseline. Annotators judged whether our model output was better, comparable, or worse than the baseline for the same input. Following the Good-Same-Bad (GSB) protocol, we compute the score as $(G-B)/(G+S+B)$.

\begin{table}[h]
  \centering
  \small
  \caption{Human evaluation using the GSB protocol. Higher is better.}
  \label{tab:gsb_single_model}
  \begin{tabular}{lc}
    \toprule
    Model & GSB Score \\
    \midrule
    FLUX.Kontext w. RL-RRM (7B) & +23.2 \\
    \bottomrule
  \end{tabular}
\end{table}

\section{Qualitative Results for FLUX-Kontext}
\label{sec:appendix_qualitative}

\begin{figure*}[h!]
    \centering
    \includegraphics[width=\textwidth]{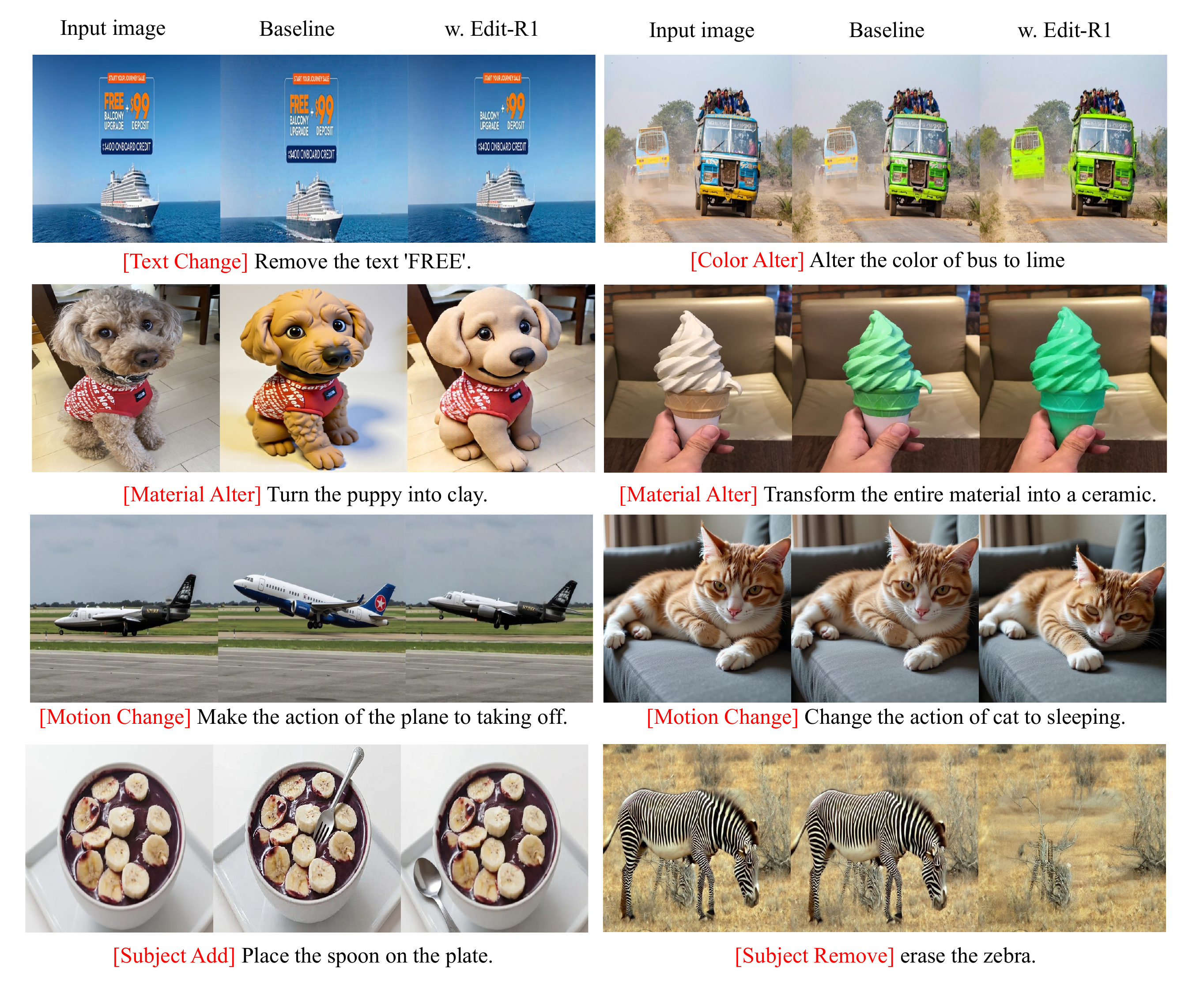} 
    \caption{
    Qualitative comparison of editing results on a diverse set of instructions. For each triplet, we show the input image, the output from the baseline model (FLUX.Kontext), and the output from our enhanced model (FLUX.Kontext w. Edit-R1). Our method demonstrates stronger performance on a broad range of editing categories, including text editing, color/material alteration, motion changes, and subject manipulation (addition and removal), producing results that better align with user instructions while maintaining high perceptual quality.
}
    \label{fig:qualitative_results}
\end{figure*}

\begin{figure*}[h!]
    \centering
    \includegraphics[width=\textwidth]{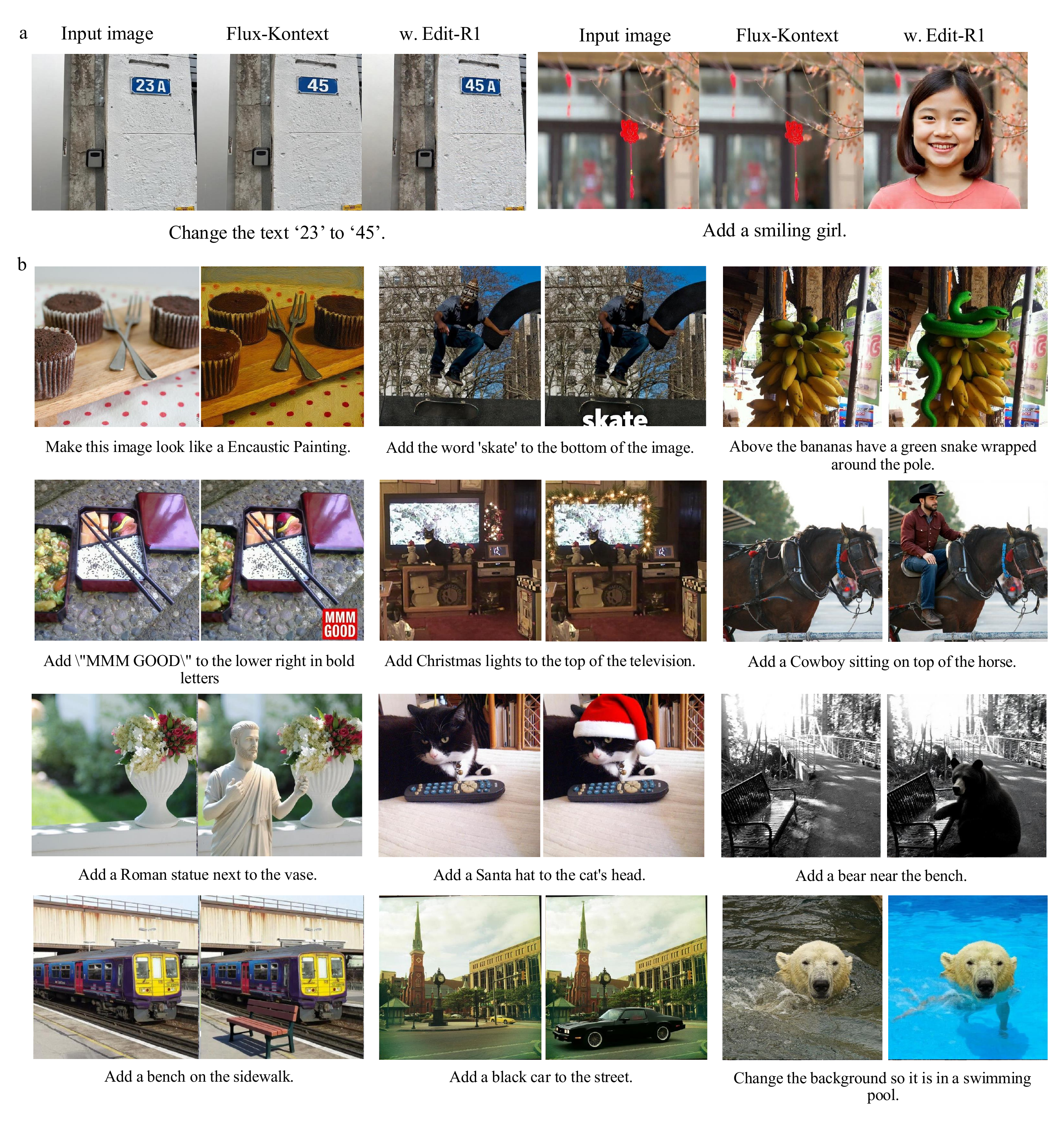} 
    \caption{
    \textbf{Further qualitative results on diverse editing benchmarks.}
    \textbf{(a)} Additional comparison results on GEdit-Bench. Our model (w. Edit-R1) consistently produces higher-quality edits that better align with user instructions compared with FLUX.Kontext baseline across tasks such as text modification and subject addition.
    \textbf{(b)} A selection of qualitative results on the challenging Emu Edit Test Set with FLUX.Kontext w. Edit-R1. These examples showcase our model's strong capability in handling a variety of complex instructions, including style transfer, object insertion with specific attributes, and substantial background alterations.
    }
    \label{fig:more_qualitative_results2}
\end{figure*}

\begin{figure*}[h!]
    \centering
    \includegraphics[width=\textwidth]{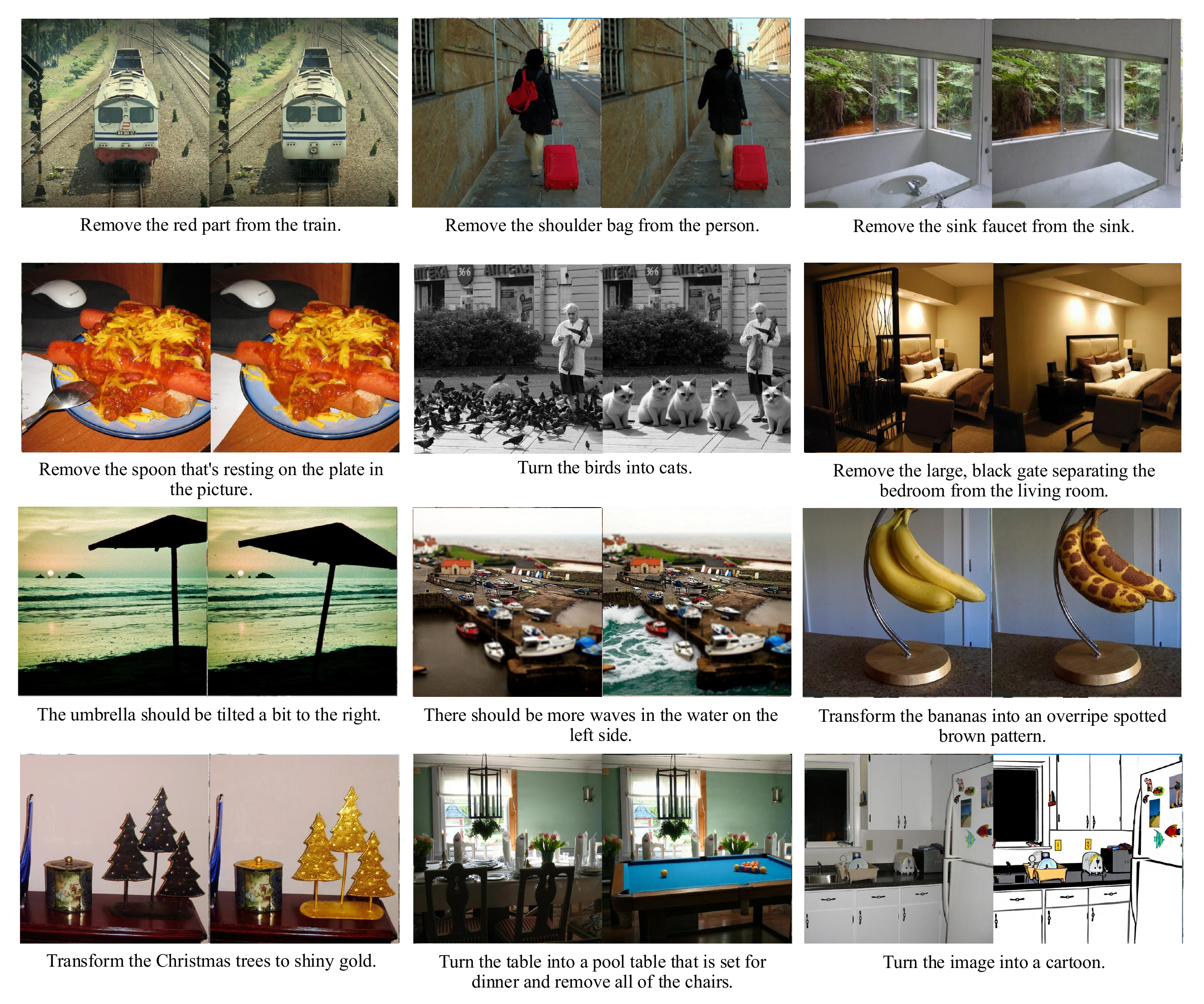} 
    \caption{
        A selection of qualitative results on the challenging Emu Edit Test Set. These examples showcase our model's robust capabilities in handling a wide variety of complex instructions.
    }
    \label{fig:more_qualitative_results3}
\end{figure*}

Qualitative results for FLUX.Kontext are shown in Fig.~\ref{fig:qualitative_results}, Fig.~\ref{fig:more_qualitative_results2}, Fig.~\ref{fig:more_qualitative_results3},.

\section{Qualitative Results for Qwen-Edit}
\label{sec:appendix_qualitative_qwenedit}
\begin{figure*}[h!]
    \centering
    \includegraphics[width=\textwidth]{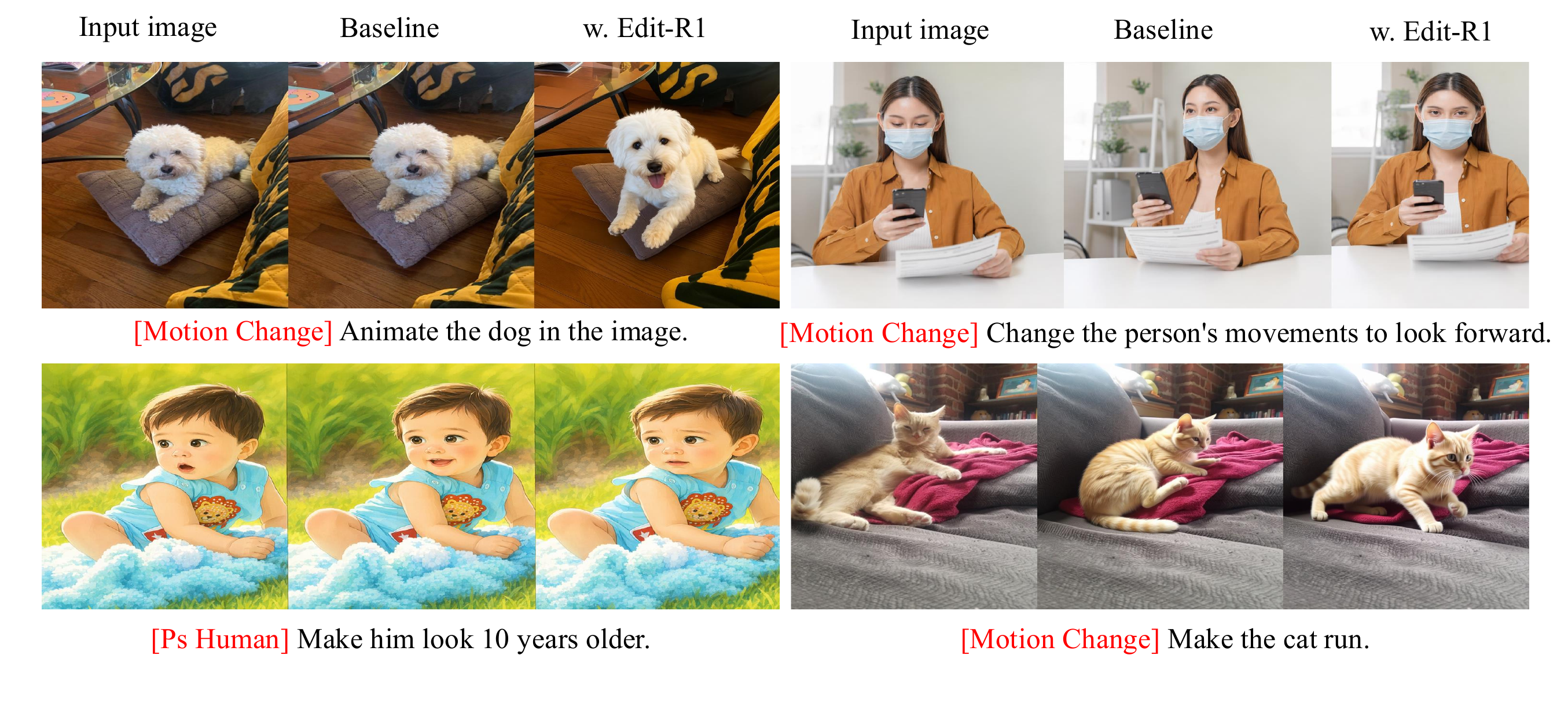} 
    \caption{
    Qualitative comparison of editing results on a diverse set of instructions. For each triplet, we show the input image, the output from the baseline model (Qwen-Edit), and the output from our enhanced model (Qwen-Edit w. Edit-R1).
    Our method further improves Qwen-Edit on challenging edits, especially motion-related edits and other fine-grained attribute changes.
    }
    \label{fig:qualitative_resultsqwen}
\end{figure*}

Qualitative results for Qwen-Edit are shown in Fig.~\ref{fig:qualitative_resultsqwen}.

\clearpage
\begin{figure*}[t!]
    \centering
    \includegraphics[width=0.7\textwidth]{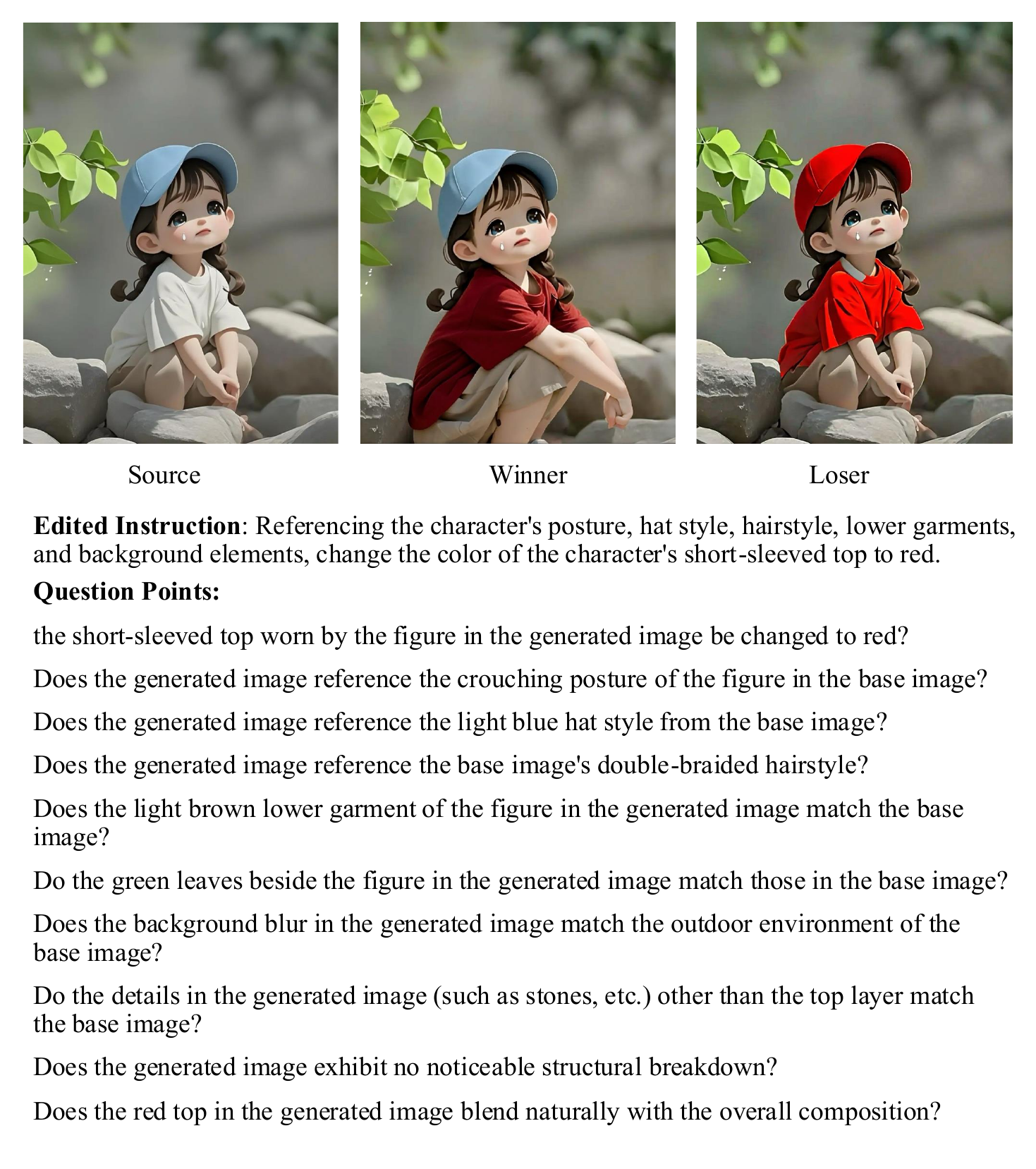} 
    \caption{
        A qualitative example illustrating how Reinforcement Learning (RL), guided by our Verifier-based Reasoning Reward Model (RRM), corrects a hallucination issue from the Supervised Fine-Tuning (SFT) model. The instruction is to change the shirt to red while preserving other features. The SFT model's "loser" output incorrectly changes the hat color to red. The RRM penalizes this failure. The RL model's "winner" output correctly preserves the blue hat, demonstrating the effectiveness of our training pipeline in resolving specific editing failures.
    }
    \label{fig:qualitative_verification}
\end{figure*}

\clearpage
\section{Qualitative Analysis of RRM Judgments}
\label{sec:appendix_qualitative_rrm_judgement}

To provide a more intuitive understanding of how our Verifier-based Reasoning Reward Model (RRM) guides the Reinforcement Learning (RL) process, this section presents a qualitative analysis of its judgments. We examine a case where the initial Supervised Fine-Tuning (SFT) model exhibits a common failure mode—hallucination—and demonstrate how the RL-tuned model, guided by the RRM's feedback, successfully corrects this error.

Fig.~\ref{fig:qualitative_verification} illustrates this process. The task is to change the color of the character's shirt to red while preserving all other features, including the light blue hat. The SFT model produces a "loser" image where it correctly changes the shirt color but incorrectly changes the hat color to red as well—a clear instance of attribute leakage or hallucination. In contrast, the RL-tuned model produces a "winner" image that accurately follows the instruction, changing only the shirt color and preserving the original blue hat.

The RRM's fine-grained evaluation is crucial here. As shown in the verification results (Listings~\ref{lst:sft_loser_verification} to \ref{lst:rl_winner_verification}), the RRM correctly identifies the SFT model's failure by assigning a score of `1` to the "loser" image for the question regarding hat style preservation (Listing~\ref{lst:sft_loser_verification}). For the RL model's "winner" image, the RRM correctly assigns a score of `1`, confirming that the hallucination was resolved (Listing~\ref{lst:rl_winner_verification}). This case study highlights the RRM's ability to provide precise, targeted feedback that enables the RL process to fix specific model weaknesses and improve instruction-following capabilities.

\begin{lstlisting}[style=promptstyle, caption={RRM verification for the SFT model's "loser" output. It correctly identifies the failure to preserve the hat style.}, label=lst:sft_loser_verification, escapechar=@]
Question: Has the short-sleeved top of the character in the generated image been changed to red?
Score: 1
--------------------
Question: Does the generated image reference the crouching posture of the character in the source image?
Score: 1
--------------------
@\textcolor{red}{\textbf{Question: Does the generated image reference the light blue hat style from the source image?}}@
@\textcolor{red}{\textbf{Score: 1}}@
--------------------
Question: Does the generated image reference the double-braided hairstyle from the source image?
Score: 1
--------------------
Question: Is the light brown lower garment of the character in the generated image consistent with the source image?
Score: 1
--------------------
Question: Are the green leaves beside the character in the generated image consistent with the source image?
Score: 1
--------------------
Question: Is the blurred outdoor background in the generated image consistent with the source image?
Score: 1
--------------------
Question: Are other details (e.g., stones) besides the top in the generated image consistent with the source image?
Score: 1
--------------------
Question: Is the generated image free of significant structural problems?
Score: 1
--------------------
Question: Does the red top in the generated image blend naturally with the overall scene?
Score: 1
\end{lstlisting}
\begin{lstlisting}[style=promptstyle, caption={RRM verification for the SFT model's "winner" output.}, label=lst:sft_winner_verification, escapechar=@]
Question: Has the short-sleeved top of the character in the generated image been changed to red?
Score: 1
--------------------
Question: Does the generated image reference the crouching posture of the character in the source image?
Score: 1
--------------------
@\textcolor{red}{\textbf{Question: Does the generated image reference the light blue hat style from the source image?}}@
@\textcolor{red}{\textbf{Score: 1}}@
--------------------
Question: Does the generated image reference the double-braided hairstyle from the source image?
Score: 1
--------------------
Question: Is the light brown lower garment of the character in the generated image consistent with the source image?
Score: 1
--------------------
Question: Are the green leaves beside the character in the generated image consistent with the source image?
Score: 1
--------------------
Question: Is the blurred outdoor background in the generated image consistent with the source image?
Score: 1
--------------------
Question: Are other details (e.g., stones) besides the top in the generated image consistent with the source image?
Score: 1
--------------------
Question: Is the generated image free of significant structural problems?
Score: 1
--------------------
Question: Does the red top in the generated image blend naturally with the overall scene?
Score: 1
\end{lstlisting}
\begin{lstlisting}[style=promptstyle, caption={RRM verification for the RL-tuned model's "loser" output. The model still fails on this specific point.}, label=lst:rl_loser_verification, escapechar=@]
Question: Has the short-sleeved top of the character in the generated image been changed to red?
Score: 1
--------------------
Question: Does the generated image reference the crouching posture of the character in the source image?
Score: 1
--------------------
@\textcolor{red}{\textbf{Question: Does the generated image reference the light blue hat style from the source image?}}@
@\textcolor{red}{\textbf{Score: 0}}@
--------------------
Question: Does the generated image reference the double-braided hairstyle from the source image?
Score: 1
--------------------
Question: Is the light brown lower garment of the character in the generated image consistent with the source image?
Score: 1
--------------------
Question: Are the green leaves beside the character in the generated image consistent with the source image?
Score: 1
--------------------
Question: Is the blurred outdoor background in the generated image consistent with the source image?
Score: 1
--------------------
Question: Are other details (e.g., stones) besides the top in the generated image consistent with the source image?
Score: 1
--------------------
Question: Is the generated image free of significant structural problems?
Score: 1
--------------------
Question: Does the red top in the generated image blend naturally with the overall scene?
Score: 1
\end{lstlisting}
\begin{lstlisting}[style=promptstyle, caption={RRM verification for the RL-tuned model's "winner" output. It confirms the model has learned to preserve the hat style correctly.}, label=lst:rl_winner_verification, escapechar=@]
Question: Has the short-sleeved top of the character in the generated image been changed to red?
Score: 1
--------------------
Question: Does the generated image reference the crouching posture of the character in the source image?
Score: 1
--------------------
@\textcolor{red}{\textbf{Question: Does the generated image reference the light blue hat style from the source image?}}@
@\textcolor{red}{\textbf{Score: 1}}@
--------------------
Question: Does the generated image reference the double-braided hairstyle from the source image?
Score: 1
--------------------
Question: Is the light brown lower garment of the character in the generated image consistent with the source image?
Score: 1
--------------------
Question: Are the green leaves beside the character in the generated image consistent with the source image?
Score: 1
--------------------
Question: Is the blurred outdoor background in the generated image consistent with the source image?
Score: 1
--------------------
Question: Are other details (e.g., stones) besides the top in the generated image consistent with the source image?
Score: 1
--------------------
Question: Is the generated image free of significant structural problems?
Score: 1
--------------------
Question: Does the red top in the generated image blend naturally with the overall scene?
Score: 1
\end{lstlisting}

\end{CJK*}
\end{document}